\documentclass[10pt, a4paper]{article}
\usepackage[a4paper,left=1.2in,top=1.25in,right=1.2in,bottom=1.5in]{geometry}

\usepackage[utf8]{inputenc} 
\usepackage[T1]{fontenc}    
\usepackage{hyperref}       
\usepackage{url}            
\usepackage{nicefrac}       
\usepackage{microtype}      
\usepackage{booktabs} 
\usepackage{times}
\usepackage{amssymb,amsmath,amsfonts}
\usepackage{graphicx}   
\usepackage{verbatim}   
\usepackage{color}      
\usepackage{stmaryrd}
\usepackage{bbm}
\usepackage{multirow}
\usepackage{pifont}
\usepackage{textcomp}
\usepackage{dsfont}
\usepackage{algorithm}
\usepackage{algorithmic}
\usepackage{mathtools}
\usepackage{subfigure}
\usepackage{array}
\usepackage[round]{natbib} 

\newtheorem{theorem}{Theorem}
\newtheorem{definition}{Definition}
\newtheorem{lemma}{Lemma}

\newtheorem{corollary}{Corollary}

\newcommand{\bp}{\noindent{\emph{Proof}.}\ }
\newcommand{\ep}{\hfill $\Box$}

\newcommand{\BEAS}{\begin{eqnarray*}}
\newcommand{\EEAS}{\end{eqnarray*}}
\newcommand{\BEA}{\begin{eqnarray}}
\newcommand{\EEA}{\end{eqnarray}}
\newcommand{\BEQ}{\begin{equation}}
\newcommand{\EEQ}{\end{equation}}
\newcommand{\BIT}{\begin{itemize}}
\newcommand{\EIT}{\end{itemize}}
\newcommand{\BNUM}{\begin{enumerate}}
\newcommand{\ENUM}{\end{enumerate}}
\newcommand{\beq}{\begin{equation}}
\newcommand{\eeq}{\end{equation}}
\newcommand{\beqa}{\begin{eqnarray}}
\newcommand{\eeqa}{\end{eqnarray}}
\newcommand{\beqan}{\begin{eqnarray*}}
\newcommand{\eeqan}{\end{eqnarray*}}
\newcommand{\bealn}{\begin{align*}}
\newcommand{\eealn}{\end{align*}}

\newcommand{\als}[1]{ \begin{align*} #1  \end{align*}}

\newcommand{\sk}{\nonumber\\}

\newcommand{\el}{\end{flushleft}}
\newcommand{\bl}{\begin{flushleft}}

\newcommand{\argsort}{\mathop{\mathrm{argsort}}}

\newcommand{\bA}{\mathbb{A}}

\newcommand{\cA}{\mathcal{A}}
\newcommand{\cB}{\mathcal{B}}

\newcommand{\cE}{\mathcal{E}}
\newcommand{\cF}{\mathcal{F}}

\newcommand{\cL}{\mathcal{L}}
\newcommand{\cM}{\mathcal{M}}

\newcommand{\cO}{\mathcal{O}}
\newcommand{\cP}{\mathcal{P}}
\newcommand{\cQ}{\mathcal{Q}}
\newcommand{\cR}{\mathcal{R}}
\newcommand{\cS}{\mathcal{S}}

\newcommand{\cX}{\mathcal{X}}

\newcommand{\kR}{\mathfrak{R}}

\newcommand{\Real}{\mathbb{R}}
\newcommand{\Nat}{\mathbb{N}}

\newcommand{\indic}[1]{\mathbb{I}\{#1\}}

\renewcommand{\phi}{\varphi}
\renewcommand{\epsilon}{\varepsilon}

\def\NN{{\mathbb N}}
\def\EE{{\mathbb E}}
\def\PP{{\mathbb P}}
\def\RR{{\mathbb R}}

\newcommand{\Esp}{\mathbb{E}}
\renewcommand{\Pr}{\mathbb{P}}

\newcommand{\UCRLBernstein}{{\text{UCRL2-B}}}
\newcommand{\UCRLSA}{\texttt{UCRL-PRM}}
\newcommand{\UCRLSAL}{\texttt{UCRL-PRM-L1}}
\newcommand{\UCRLSAB}{\texttt{UCRL-PRM-B}}
\newcommand{\UCRL}{\text{UCRL2}}

\newcommand{\E}{\mathbb{E}}
\newcommand{\eventFunction}{L}
\newcommand{\Dcp}{D^\times} 
\newcommand{\Dcpsq}{D^{\times 2}} 
\newcommand{\EVI}{\texttt{EVI}}
\newcommand{\cpM}{M^\times} 
\newcommand{\cpP}{P^\times} 
\newcommand{\cpR}{R^\times} 

\newcommand{\AtomProp}{\textsf{AP}}
\newcommand{\counter}{n}

\renewcommand{\indic}[1]{\mathbb I_{\{#1\}}}

\usepackage[colorinlistoftodos, textwidth=4cm, shadow]{todonotes}

\newcommand{\vertiii}[1]{{\left\vert\kern-0.25ex\left\vert\kern-0.25ex\left\vert #1
    \right\vert\kern-0.25ex\right\vert\kern-0.25ex\right\vert}}

\title{Provably Efficient Exploration in Reward Machines\\ with Low Regret}
\author{Hippolyte Bourel\thanks{Department of Computer Science,
University of Copenhagen}, Anders Jonsson\thanks{Department of Information and Communication Technologies, Universitat Pompeu Fabra}, Odalric-Ambrym Maillard\thanks{University of Lille, Inria, CNRS, Centrale Lille}\\ Chenxiao Ma\thanks{Department of Computer Science, University of Copenhagen}, and Mohammad Sadegh Talebi\thanks{Department of Computer Science,
University of Copenhagen. Corresponding author (email: \texttt{sadegh.talebi@di.ku.dk}).}} 
\date{\today}

\begin{document}

\maketitle

\begin{abstract}
    We study reinforcement learning (RL) for decision processes with non-Markovian reward, in which high-level knowledge of the task in the form of reward machines is available to the learner. We consider probabilistic reward machines with initially unknown dynamics, and investigate RL under the average-reward criterion, where the learning performance is assessed through the notion of regret. Our main algorithmic contribution is a model-based RL algorithm for decision processes involving probabilistic reward machines that is capable of exploiting the structure induced by such machines. We further derive high-probability and non-asymptotic bounds on its regret and demonstrate the gain in terms of regret over existing algorithms that could be applied, but obliviously to the structure. 
We also present a regret lower bound for the studied setting. To the best of our knowledge, the proposed algorithm constitutes the first attempt to tailor and analyze regret specifically for RL with probabilistic reward machines.
\end{abstract}

\section{Introduction}\label{sec:intro}

Most state-of-the-art reinforcement learning (RL) algorithms assume that the underlying decision process has Markovian reward and dynamics, i.e.~that future observations depend only on the current state-action of the system. In this case, the Markov Decision Process (MDP) is a suitable mathematical model to represent the task to be solved \citep{puterman2014markov}. However, there are many application scenarios with non-Markovian reward and/or dynamics~\citep{bacchus1996rewarding,brafman2019regular,littman2017environment} that are more appropriately modeled as \emph{Non-Markovian Decision Processes (NMDPs)}. 

NMDPs capture environments in which optimal action depends on events that occurred in the past, implying that the learning agent has to remember parts of the history. For example, a robot may receive a reward for delivering an item only if the item was requested previously, and a self-driving car is more likely to skid and lose control if it rained previously. Consider a mobile robot that has to track an object which is no longer in the robot's field of view. By remembering where the object was last seen, the robot has a better chance of discovering the object again. An even more precise estimation is given by the sequence of last observations (which also capture direction of movement). This can be formalized by defining high-level events that correspond to past observations. 

In general, the future observations of an NMDP can depend on an infinite history or trace, preventing efficient learning. Consequently, recent research has focused on tractable sub-classes of NMDPs. A tractable and recently introduced sub-class is Regular Decision Processes (RDPs)~\citep{brafman2019regular,brafman2024regular}, where the reward function and next state distribution is determined by conditions over the history that fall within the class of the regular languages. Another popular formalism enjoying tractability is the {\em Reward Machine} (RM)~\citep{icarte2018using,icarte2022reward}, which is a Deterministic Finite-State Automaton (DFA) providing a compact representation of history that compresses the entire sequence of past events into a single state, which can be combined with the current observation to determine the best action. Hence, the current state of the reward machine is sufficient to fully specify the reward function. Nevertheless, high-level deterministic transitions often fall short of representing real-world cases. Keys may need to be turned several times to open a door; swiping a card may require multiple tries to succeed. To remedy such limitations, \cite{dohmen2022inferring} introduced the notion of probabilistic RMs, which we adopt in this paper. 

In this paper, we investigate RL in Markov decision processes with reward machines (MDPRMs) under the average-reward criterion, where the agent performance is measured through the notion of regret with respect to an oracle aware of the transition dynamics and associated reward functions. The goal of the  agent is to minimize its regret, which entails balancing exploration and exploitation. We focus on setting where dynamics of both observations and RM states are \textit{unknown}, while states are observable. For a given MDPRM, it is possible to formulate an \emph{equivalent cross-product} MDP (adhering to the Markov property) as discussed in the literature \citep{icarte2018using} and apply provably efficient off-the-shelf algorithms \emph{obliviously} to the structure induced by the MDPRM. However, this would lead to large regret, both empirically and theoretically, as the associated cross-product MDP usually has a large state-space. Therefore, sample-efficient learning of near-optimal policies entails exploiting the intrinsic structure of MDPRMs in an efficient manner.

\subsection{Contributions}
We make the following contributions. We formalize regret minimization in MDPRMs with probabilistic machines under the average-reward criterion, and establish a first, to the best of our knowledge, regret lower bound for MDPRMs. We introduce a provably efficient algorithm called \UCRLSA, which implements the principle of \emph{optimism in the face of uncertainty} through a model-based approach, whose design is inspired by the celebrated \UCRL\ algorithm \citep{jaksch2010near}, which guarantees a near-optimal regret bounds in the class of communicating MDPs without any prior knowledge on the MDP. 
\UCRLSA\ uses high-probability and \emph{time-uniform} confidence sets for unknown parameters of the underlying MDPRM and performs policy optimization over the correspondingly defined set of plausible MDPRMs. However, \UCRLSA\ is carefully tailored to leverage the structure in MDPRMs, which is a key departure from \UCRL-style algorithms for MDPs. Specifically, we derive two variants of \UCRLSA\ that mainly differ in the choice of confidence sets used: \UCRLSAL, which uses $L_1$-type confidence sets, and \UCRLSAB\, relying on Bernstein concentration. As a result, they attain different regret bounds.

More precisely, we show that \UCRLSAL\ achieves a high-probability regret growing as $\widetilde \cO\big(\Dcp\sqrt{(Q^2E + O^2A)T}\big)$ in an MDPRM $M$ after $T$ steps of interaction, with $O$ and $A$ being the respective size of observation and action spaces, $Q$ denoting the number of states of the RM, and $E$ denoting the maximum number of relevant labels at any RM state.\footnote{The notation $\widetilde O(\cdot)$ hides poly-logarithmic terms in $T$ and numerical constants.} Finally, $\Dcp$ denotes the diameter of the cross-product MDP associated to $M$. In the case of \UCRLSAB, we derive a regret bound informally growing as  
$\widetilde \cO\big(\Dcp\sqrt{(OAK + QEK')T}\big)$, where $K$ and $K'$ are the respective support size of the next-state of observations and RM states. These bounds improve over the regret bounds of baselines that scale as $\widetilde\cO\big(\Dcp QO\sqrt{AT}\big)$ and $\widetilde\cO\big(\Dcp \sqrt{QOAK''T}\big)$ (for some $K''\ge \max\{K,K'\}$).\footnote{For further details, we refer to Section \ref{sec:regret_bound}.}  

In addition, we establish refined regret bounds for \UCRLSA\ in the case of deterministic machines. Specifically, we show that in an MDPRM $M$ with deterministic RM, \UCRLSAL\ (resp.~\UCRLSAB) achieves a regret growing as $\widetilde O(\sqrt{\mathbf c_M OAT})$ (resp.~$\widetilde O(\sqrt{\mathbf c'_M OAT})$), where $\mathbf c_M$ and $\mathbf c'_M$ are problem-dependent quantities defined in terms of a novel notion of connectivity in MDPRMs, which we call the \emph{RM-restricted diameter}. This notion is a problem-dependent refinement of the diameter $\Dcp$ of the cross-product MDP associated to $M$. The RM-restricted diameter of $M$ reflects the connectivity in $M$ \emph{jointly} determined by the dynamics and the sparsity structure of the RM, and we believe it could be of interest in other settings of reward machines. The RM-restricted diameter is \emph{always} smaller than $\Dcp$, and in some MDPRM instances, it is proportional to $\Dcp/Q$. 

Although the design and analysis of \UCRLSA\ build on UCRL2 and its variants, we stress that there are some non-trivial components.  First, directly using the policy optimization procedure of UCRL2-style algorithms would require solving a bilinear program, which is NP-hard in general. To circumvent this issue, we perform policy optimization over a surrogate set of candidate MDPRMs, which entails solving a linear program. Second, the analysis of \UCRLSA\ tackle the structural properties of MDPRMs, which in turn leads to making appear the RM-restricted diameters in the regret bound. 

This paper builds upon our previous work \citep{bourel2023exploration}, where we originally investigated RL in average-reward MDPRMs in the regret setting. It restricted attention to the case of deterministic RMs, and proposed and analyzed \texttt{UCRL-RM}. We extend the setting of \citep{bourel2023exploration} to the case of MDPRMs with probabilistic RMs by presenting \UCRLSA\ and analyzing its high-probability regret. In terms of algorithmic novelty, the present work crucially relies on carefully chosen surrogate set of MDPRMs to perform policy computation. Further,  
the regret analysis of \UCRLSA\ renders more challenging than deterministic RMs. To the best of our knowledge, the present paper and its preceding work \citep{bourel2023exploration} are the first studying regret minimization in average-reward MDPRMs, and the proposed algorithms constitute the first attempt to tailor and analyze regret specifically for MDPRMs or MDPs with associated DFAs.


\subsection{Related Work}
In the case of Markovian rewards and dynamics, there is a rich literature on average-reward RL, presenting several algorithms with theoretical regret guarantees. While there is an abundance of work on the tabular case (i.e., without structural assumptions), there is a well-growing line of work on structured RL. For the former category we mention   \citep{burnetas1997optimal,jaksch2010near,fruit2018near,talebi2018variance,wei2020model,bourel2020tightening,zhang2019regret,pesquerel2022imed,saber2023logarithmic}), whereas some the latter one include \citep{wei2021learning,ok2018exploration,talebi2021improved,lakshmanan2015improved}. 
In the absence of structure assumptions, as established by \citet{jaksch2010near}, no algorithm can have a regret lower than $\Omega(\sqrt{DSAT})$ in a communicating MDP with $S$ states, $A$ actions, diameter $D$, and after $T$ steps of interactions. The best available regret bounds for communicating MDPs, achievable by computationally implementable algorithms, grow as $\cO(\sqrt{DSAKT}\log(T))$ \citep{fruit2020improved} or as $\cO(D\sqrt{KSAT\log(T)})$ \citep{fruit2018near}, where $K$ denotes the maximal number of next-states under any state-action pair in the MDP. 
Recently, \citet{boone2024achieving} present an algorithm achieving a regret of $\cO(\sqrt{DSAT\log(T)})$, albeit with an additive term scaling as $S^{5/2}T^{9/20}$ making the bound less interesting. 


The progress in the domain of non-Markov RL has been substantially slower than the Markovian counterpart due to challenges posed by history dependence. A generic NMDP, with rewards and transition function arbitrarily depending on the history, is not PAC learnable. 
Nevertheless, there is already a broad literature in various sub-classes of NDMPs that admit some form of tractability, making learning a possibility. 
A line of such work tackle the problem of state-representation, where the agent must select a representation of the environment (i.e., a mapping from histories to a discrete state-space) from an input set~\citep{lattimore2013general,maillard2013repr,sunehag2015general}. Although these algorithms could be applied to RMs, they do not exploit the particular structure of the underlying RMs, and hence the resulting theoretical bounds may grow prohibitively large. As a result, state representation learning algorithms render impractical for learning RMs. 

RDPs \citep{brafman2024regular} constitute another tractable class of NMDPs, which can be modeled via some unobservable DFA. More precisely, the automaton state of an RDP may be viewed as a  hidden information state \citep{subramanian2022approximate}, and as shown by \cite{brafman2024regular}, any RDP is also a Partially-Observable Markov Decision Process (POMDP) \citep{kaelbling1998pomdps}, whose hidden dynamics evolve according to its finite-state automaton. As a result, RMs may fall into the class of RDPs; however, they are simpler to learn because of full observability. RL in RDPs is a recent under-taking and remains mostly unexplored. The S3M algorithm of \citet{abadi2020learning} integrates RL with the logical formulas of RDPs, but does not admit polynomial sample complexity in the PAC setting. \citet{ronca2021efficient} present the first online RL algorithm for RDPs whose PAC sample complexity grows polynomially in terms of the underlying parameters, though the sample complexity bound is prohibitively large in the relevant parameter. Recently, \cite{cipollone2024provably} introduced RegORL, a provably efficient algorithm for offline RL in RDPs with near-optimal sample complexity. Nevertheless, none of these work could be used to control exploration in RMs with provable regret guarantees. 

Research on reward machines is relatively recent, but has grown quickly in popularity and already attracted many researchers to the field. Initial research focused on proving convergence guarantees for RL algorithms specifically devised for RMs~\citep{icarte2018using,icarte2022reward}. There is also a rich literature on RL with temporal specifications expressed in Linear Temporal Logic (LTL)~\citep{camacho2019ltl,kazemi2022translating}. Because of the equivalence between LTL and B\"uchi automata, LTL specifications are often translated to DFAs similar to RMs, and sometimes combined with hierarchical RL~\citep{denhengst2022reinforcement}. More recently, many researchers have investigated how to learn RMs or similar DFAs from experience in the form of traces~\citep{abate2022learning,giacomo2020bolts,furelos2021induction,gaon2020reinforcement,toro2019learning,verginis2022joint,xu2020inference}, and extensions to stochastic and probabilistic RMs in which either the rewards or the transitions are non-deterministic exist~\citep{corazza2022reinforcement,dohmen2022inferring}. Another recent extension is to learn entire hierarchies of RMs \citep{furelos2023hierarchies}. RMs have also been used in combination with multiagent RL~\citep{dann2022multi,neary2021reward}. 
Among the fast growing literature on RMs literature, little attention is paid to regret minimization. This paper builds on our previous work \citep{bourel2023exploration}, which is the first, to our knowledge, studying average-reward RL in the regret setting in RMs with deterministic dynamics.  We are not aware of any other work involving RMs that report regret bounds in the average-reward setting. The only available work in the episodic setting is due to \cite{lin2024efficient}, who study regret in episodic RL in probabilistic RMs but assuming known RM dynamics. Hence, the machinery used \citep{lin2024efficient} does not apply to our case. 

\subsection{Organization}

The remaining of the paper is organized as follows. In Section \ref{sec:pb_formulation}, we introduce the notion of MDPRM under the average-reward criterion and formulate the corresponding regret minimization problem. We present two variants of the \UCRLSA\ algorithm in Section \ref{sec:algos}, and report high-probability and finite-time bounds on their regret in Section \ref{sec:regret_bound}. A regret lower bound for MDPRMs with deterministic machines is derived in Section \ref{sec:regret_LB}. Finally, Section \ref{sec:conclusion} concludes the paper and provides some some future research directions. Proofs as well as some algorithmic details are presented in the appendix.

\section{Problem Formulation}\label{sec:pb_formulation}

\subsection{MDPRMs: Average-Reward Markov Decision Processes with Reward Machines}
We begin with introducing some necessary background on labeled Markov decision processes and reward machines. We introduce notations that will be used throughout. Given a set $A$, $\Delta_A$ denotes the simplex of probability distributions over $A$. 
$A^*$ denotes (possibly empty) sequences of elements from $A$, and $A^+$ denotes non-empty sequences. $\mathbb{I}_{A}$ denotes the indicator function of event $A$.

\subsubsection{Labeled Markov Decision Processes}A \emph{labeled average-reward MDP} is a tuple $M = (\cO,\cA,P,\mathbf R, {\AtomProp}, L)$, where $\cO$ is a finite set of (observation) states with cardinality $O$, $\cA$ is a finite set of actions available at each state with cardinality $A$, $P:\cO\!\times\!\cA\to \Delta_\cO$ is the transition function such that $P(o'|o,a)$ denotes the probability of transiting to state $o'\in \cO$, when executing action $a\in \cA$ in state $o\in \cO$. $\mathbf R:(\cO\!\times\!\cA)^+\to\Delta_{[0,1]}$ denotes a history-dependent reward function such that for every history $h\in (\cO\!\times\!\cA)^*\times\cO$ and action $a\in \cA$, $\mathbf R(h,a)$ defines a reward distribution.
\footnote{
This can be straightforwardly extended to $\sigma$-sub-Gaussian reward distributions with unbounded supports.}  
Further, ${\AtomProp}$ denotes a set of atomic propositions and $L\!:\!\cO\!\times\!\cA\!\to\!2^{\AtomProp}$ denotes a labeling function assigning a subset of ${\AtomProp}$ to each $(o,a)$.\footnote{A more complex labeling function of the form $L\!:\!\cO\!\times\!\cA\!\times\!\cO\!\to\!2^{\AtomProp}$ could be considered.} 
These labels describe high-level events associated to the various $(o,a)$ pairs that can be detected from the environment, and they prove instrumental in defining the history-dependent reward function $\mathbf R$.

The notion of $M$ above coincides with the conventional notion of average-reward MDPs except that (i) it assumes a \emph{non-Markovian} reward function and (ii) it is equipped with a labeling mechanism  (defined via $L$ and $\AtomProp$).
The interaction between the agent and the environment $M$ proceeds as follows. 
At each time step $t\in \mathbb N$, the agent is in state $o_t\in \cO$ and chooses an action $a_t\!\in\!\cA$ based on $h_t\!:=\!(o_1,a_1,\ldots,o_{t-1},a_{t-1},o_t)$. Upon executing $a_t$ in $o_t$, $M$ generates a next-state $o_{t+1}\sim P(\cdot|o_t,a_t)$ and assigns a label $\sigma_t\!=\!L(o_t,a_t)$. 
Then, the agent receives a reward $r_t\sim\mathbf R(h_t, a_t)$. 
Then, the state transits to $o_{t+1}$  and a new decision step begins. As in MDPs, after $T$ steps of interactions, the agent's  cumulative reward is  $\sum_{t=1}^T r_t$. 

\subsubsection{(Probabilistic) Reward Machines}We restrict attention to a class of non-Markovian reward functions that are encoded by RMs \citep{icarte2022reward,dohmen2022inferring}, whose definitions are inspired by conventional 
DFAs. In this work, we consider probabilistic RMs \citep{dohmen2022inferring}. A probabilistic RM is a tuple $\mathcal{R} = (\mathcal{Q}, 2^{\AtomProp}, \tau, \nu)$, where $\mathcal{Q}$ is a finite set of states and $2^{\AtomProp}$ is an input alphabet. 
$\tau: \mathcal{Q}\!\times\!2^{\AtomProp}\!\rightarrow\!\Delta_\mathcal{Q}$ denotes a transition function such that $\tau(q'|q,\sigma)$ denotes the probability that $\cR$ transits to $q'\in \cQ$  when an input $\sigma$ is received in state $q$, with the convention that $\tau(q,\emptyset)\!=\!q$. Finally, $\nu\!:\! \mathcal{Q}\!\times\!2^{\AtomProp}\!\rightarrow\!\Delta_{[0,1]}$ denotes the output function of $\cR$, which returns a distribution over $[0,1]$ for any $(q,\sigma)$.\footnote{This is very similar to the standard definition of RM by  \citet{icarte2022reward}, though in our case the set of terminal states is empty. 
} 
Let $\cE_q$ be the set of relevant labels at $q\in \cQ$ and $E_q$ be its cardinality. Further, define $E:=\max_{q} E_q$. Note the labeling function is not necessarily one-to-one; i.e., there might exist two or more observation-action pairs generating the same label. 

In the case of deterministic transition function $\tau:\cQ\times 2^{\AtomProp}\to\cQ$, $\cR$ coincides with the conventional notion of RM considered in \cite{icarte2022reward}. In this paper, we use RM to refer to both deterministic and probabilistic machines. 
In words, the RM $\cR$ converts a (sequentially received) sequence of labels to a sequence of Markovian reward distributions such that the output reward function at time $t$ is 
$\nu(q_t,\sigma_t)$, and it thus only depends on the current state $q_t$ and current label $\sigma_t$. Conditioned on $(q_t,\sigma_t)$, 
the reward distribution at time $t$ is independent of earlier labels and RM states 
$(q_1,\sigma_1,\ldots,q_{t-1},\sigma_{t-1})$. Thus, RMs provide a compact representation for a class of non-Markovian rewards that can depend on the entire history. 


\subsubsection{Average-Reward MDPs with RMs}
\begin{figure}
    \centering
    \includegraphics[width=0.45\linewidth]{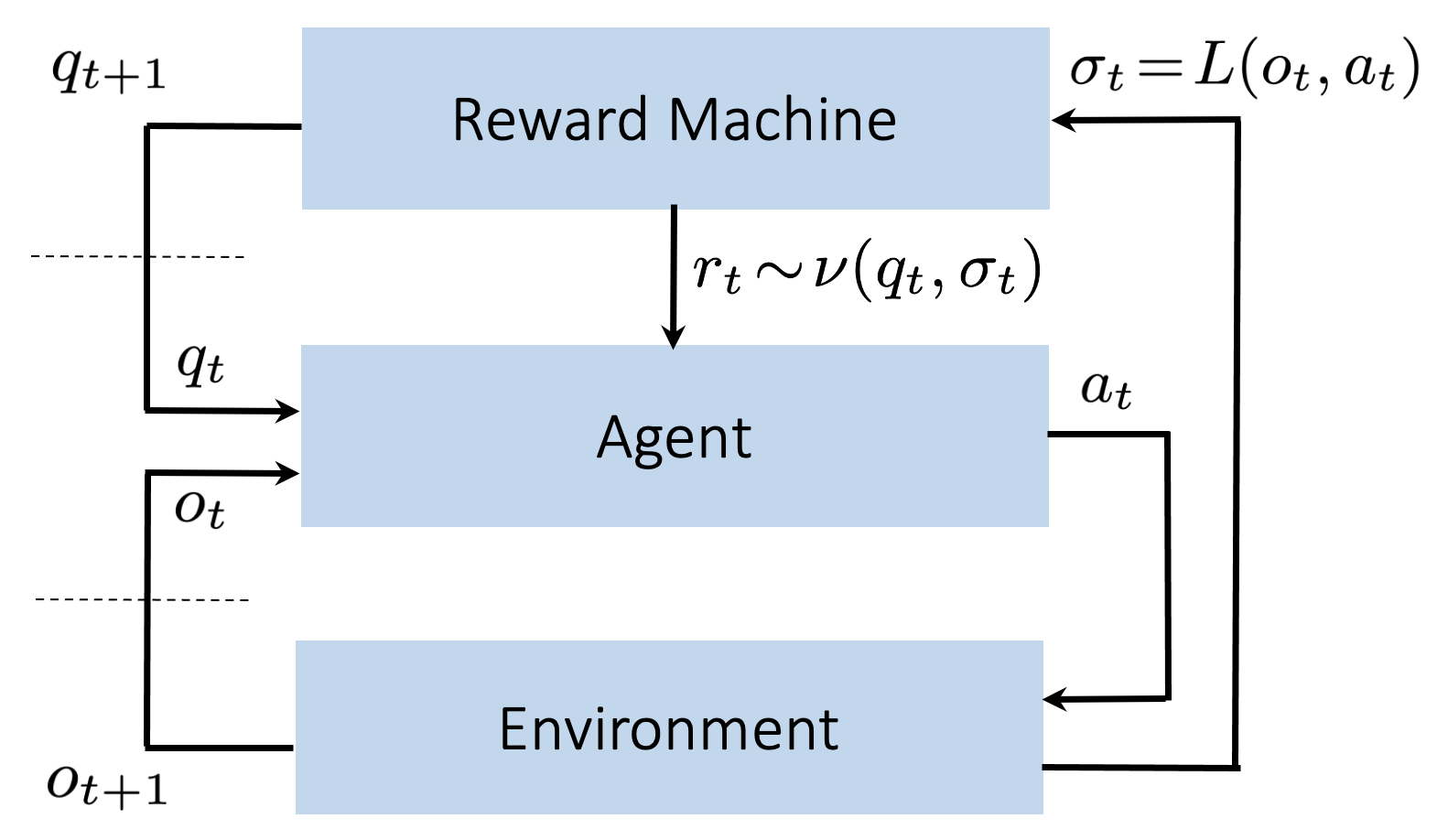}
    \caption{Interaction with an MDPRM}
    \label{fig:MDPRM_protocol}
\end{figure}

Restricting the generic history-dependent reward function $\mathbf R$ to RMs leads to decision processes that are often termed MDPRMs. Formally, an average-reward MDPRM is a tuple $M\!=\!(\cO,\cA,P,\cR,\AtomProp,L)$, where $\cO,\cA,P,{\AtomProp}$, and $L$ are defined as in (labeled) average-reward MDPs, and where $\cR$ is an RM, which generates reward functions. The agent's interaction with an MDPRM $M$ proceeds as follows. At each time $t\in \mathbb N$, the agent observes both $o_t\in\cO$ and $q_t\in\cQ$, and chooses an action $a_t\in\cA$ based on $o_t$ and $q_t$ as well as (potentially) her past decisions and observations. The environment reveals an event $\sigma_t=\eventFunction(o_t,a_t)$. The RM $\cR$,  being in state $q_t$, receives $\sigma_t$ and outputs a reward distribution 
$\nu(q_t, \sigma_t) \in \Delta_{[0,1]}$. Then, the agent receives a reward 
$r_t\sim \nu(q_t,\sigma_t)$ (at the end of the current time step). Then, the environment and RM states transit to their next states $o_{t+1}\sim P(\cdot|o_t,a_t)$ and $q_{t+1} \sim \tau(\cdot|q_t,\sigma_t)$, and a new step begins. This is summarized in Figure \ref{fig:MDPRM_protocol}.

\begin{figure}[h]
\centering
\tiny
  {%
    \subfigure[The laundry gridworld environment]{\label{fig:room}%
      \def\svgwidth{0.5\linewidth}
      \includegraphics[width=0.35\linewidth]{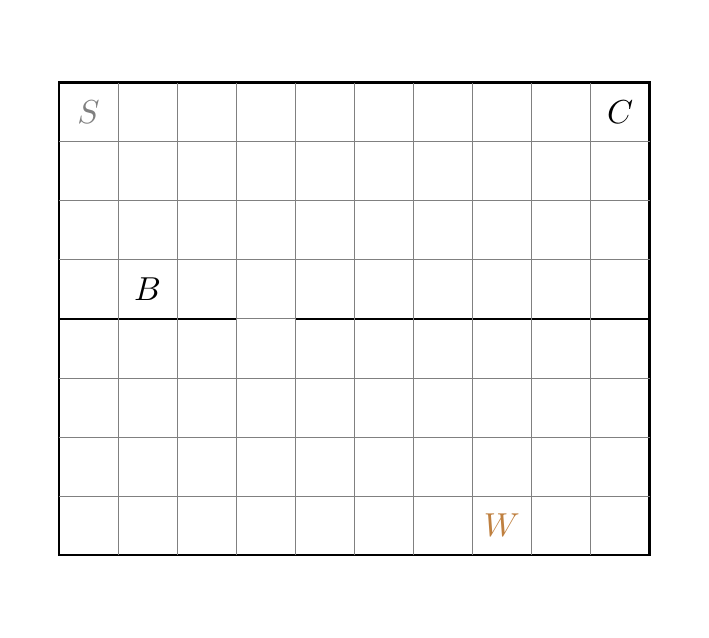}}
      \hspace{6mm}
    \subfigure[laundry RM]{\label{fig:prm_states}%
      \def\svgwidth{0.35\linewidth}
      \includegraphics[width=0.45\linewidth]{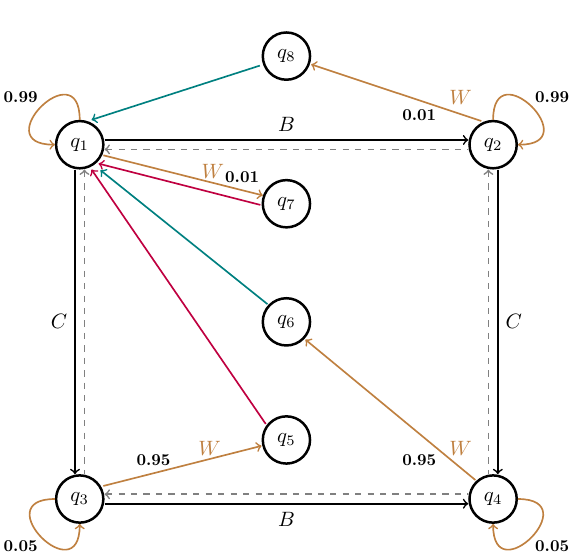}}
  }
 \vspace{-1mm}
  \caption{An example environment with one RM.}
  \label{fig:prm_env}
\vspace{-2mm}
\end{figure}

Figure \ref{fig:prm_env} illustrates an example MDPRM, which we shall call \emph{laundry gridworld}. In this example, the task consists in transferring soiled garments from a basket $B$, situated in the upper hall, to a washing machine $W$, located in the lower hall. An access card must be collected from location $C$ to use laundry machine $W$. Upon completion of the assigned task and return of the card and basket to their original locations, the agent will receive a reward. However, if the agent operates the machine without clothes of laundry, a penalty will be imposed. The environment is illustrated in Figure \ref{fig:room}. Figure \ref{fig:prm_states} depicts the corresponding RM, where the green arrows indicate the available pathways for obtaining the reward, while the red one corresponds to the misoperation. In particular, in contrast to the high probability of $0.95$ associated with operating the machine $W$ with a card, the agent can only directly operate it with a probability of (w.p.) $0.01$. 
Observation dynamics are defined similarly to a classical gridworld. Specifically, in each state, the agent has the four cardinal actions, corresponding to movement in the up, down, right, and left directions, but with uncertain transitions. A specific action moves the agent in the intended direction with a probability of $0.7$, while it may result in a movement to each perpendicular direction w.p.~$0.15$. In addition, the walls will be treated as reflectors, thereby keeping the agent in a same state. We remark that the current MDP observation (i.e., location) is not sufficient to predict what to do next, and therefore has to be combined with the current RM state.

As Figure \ref{fig:prm_states} depicts, there exist two distinct but equally desirable ways to enter the lower hall. The agent may either collect the card prior to bringing the basket or vice versa, depending on the route taken, which is indicated by the arrival to $q_4$ from $q_1$ via $q_2$ or $q_3$, respectively. Likewise, RM states may not define the optimal solution \emph{per se}, so a combination with MDP states is expected. As for the suboptimal cases, holding a card increases the probability of operating the machine, but not having a basket results in penalties, represented by the transitions from $q_3$ to $q_1$ via $q_5$. This also applies to the bare-handed agent, denoted by the transitions between $q_{1}$ and $q_7$. However, her chance of using the machine is slimmer, given that there is no card in her hand. As with the bare-handed agent, the agent with only one basket can also be trapped when she focuses on turning on the machine. This is captured by the transition from $q_2$ to $q_{8}$.

Despite the intricate dynamics manifesting even within a toy model, for a given  MDPRM, one can derive an equivalent tabular MDP (i.e., with a Markovian reward function) defined over state-space $\cS\!:=\!\cQ\times\!\cO$. 
As a result, this associated MDP, which we shall denote by $\cpM$, is often called the \emph{cross-product MDP} associated to $M$. The following lemma characterizes $\cpM$. Variants of  this result appeared in, e.g., \citep{icarte2022reward,dohmen2022inferring}; the following version slightly extends it to hold for reward distributions. 
\begin{lemma}\label{lem:cross_product}
Let $M\!=\!(\cO,\cA,P,\cR,\emph{\AtomProp},L)$ be a finite MDPRM. Then, an associated cross-product MDP to $M$ is $\cpM\!=\!(\cS,\cA,\cpP,\cpR)$, with $\cS=\cQ\times\cO$, where for $s\!=\!(q,o), s'\!=\!(q',o')\!\in\!\cS$, and $a\!\in\!\cA$,
\begin{align}
\label{eq:Mcp_PRM}
\cpP(s'|s,a) = P(o'|o, a)\tau(q'|q,\eventFunction(o, a))\,, \quad 
\cpR(s,a) = \nu(q,\eventFunction(o,a)). 
\end{align}
\end{lemma}

\subsection{Regret Minimization in MDPRMs}

We are now ready to formalize RL in MDPRMs in the regret minimization setting, which is the main focus of this paper. As in tabular RL, it involves an agent who is seeking to maximize its cumulative reward, and its performance is measured in terms of regret with respect to an oracle algorithm who knows and always applies a gain-optimal policy. To formally define regret, we introduce some necessary concepts. A stationary deterministic policy in an MDPRM $M$ is a mapping $\pi:\cS\!\to\!\cA$ prescribing an action $\pi(q, o)\!\in\!\cA$ for all $(q, o)\!\in\!\cS$. Let $\Pi$ be  the set of all 
such policies in $M$. The long-term average-reward (or gain) of policy $\pi\in \Pi$, when starting in $(q,o)$, is defined as:
$$
g^\pi(q,o) = \liminf_{T \rightarrow \infty}\frac{1}{T}\mathbb{E}_\pi \bigg[\sum_{t = 1}^T r_t\Big|q_1 = q, o_1=o\bigg]
$$
where $r_t\!\sim\!\mathbf r_t\!=\!\nu \big(q_t, \eventFunction(o_t, \pi(q_t,o_t))\big)$ for all $t$. Here the expectation is taken with respect to randomness in $r_t$ and over all possible histories $h_t$ (which implicitly depend on generated labels too). Let $g^\star(s) = \max_{\pi} g^\pi(s)$ denote the optimal gain over all (possibly history dependent) policies, where $s$ denotes the starting state. Any policy achieving $g^\star$ is an optimal policy. Following the same arguments as in tabular MDPs together with the equivalence between $M$ and its $\cpM$ (Lemma \ref{lem:cross_product}), it is guaranteed that there exists at least one optimal policy in $\Pi$. We restrict attention to the class of MDPRMs, whose associated cross-product MDPs are communicating,\footnote{We recall that a tabular MDP is communicating if  it is possible to reach any state from any other state under some stationary deterministic policy \citep{puterman2014markov}. Alternatively, an MDP is communicating if and only if its diameter is finite \citep{jaksch2010near}.} for which the optimal gain is independent of the starting state \citep{puterman2014markov}. 

We assume that agent observes both the RM state $q_t$ and observation state $o_t$ at each time step $t$, but is unaware of their underlying transition functions (i.e., $P$ and $\tau$). The agent interacts with $M$ for $T$ steps according to the protocol specified earlier. We define the regret of an agent (or learning algorithm) $\bA$  as 
$$
\kR(\bA,T) := Tg^\star - \sum_{t=1}^T r_t.
$$
Alternatively, the agent's objective is to minimize regret, which entails balancing exploration and exploitation. 
We stress that regret $\kR(\bA,T)$ compares the $T$-step reward collected by $\bA$ against an oracle that uses the \emph{same reward machine $\cR$} as the agent.

\section{Learning Algorithms for MDPRMs}\label{sec:algos}

In this section, we present algorithms for learning in MDPRMs that fall into the category of model-based algorithms. 
In general, the equivalence between MDPRM $M$ and its associated cross-product MDP $\cpM$ implies that one could apply any off-the-shelf algorithm to $\cpM$, as it perfectly adheres to the Markovian property. This implies that provably efficient algorithms designed for average-reward MDPs such as \UCRL\ \citep{jaksch2010near} and its variants (e.g., \citep{fruit2020improved,fruit2018efficient,bourel2020tightening}) could be directly applied to $\cpM$ while guaranteeing sublinear regret performance. However, this may lead to inefficient exploration, and thus large regret, since these generic algorithms are oblivious to the special structure of $\cpM$. Nevertheless, characterization of $\cpM$ can indeed be used as a proxy to develop learning algorithms for MDPRM. 

To simplify exposition, we assume that the reward distributions $\nu$ of the RM are known. This assumption can be easily relaxed at the expense of a slightly increased regret. We discuss in 
Appendix \ref{app:further_algo_detail} how to tailor the algorithms to the case of unknown rewards. 

\subsection{Confidence Sets}
We begin with introducing empirical estimates and confidence sets used by the algorithms. 
We first present confidence sets for observation dynamics $P$ and RM dynamics $\tau$, and then show how they yield confidence sets for the transition and reward functions of the cross-product MDP $\cpM$. 

\subsubsection{Confidence Sets for Observation Dynamics $P$}
Formally, under a given algorithm and for any $o,o'\in \cO$ and $a\in \cA$, let $N_t(o,a,o')$ denote the number of times a visit to $(o, a)$ was followed by a visit to $o'$, up to time $t$: $N_t(o, a,o') := \sum_{i=1}^{t-1}\mathbb I_{\{(o_{i},a_{i},o'_{i+1})=(o,a,o')\}}$. Further, $N_t(o, a) := \max\{1, \sum_{o'}N_t(o,a, o')\}$. Using the observations collected up to $t\ge 1$, we define the empirical estimate $\widehat P_t(o'|o,a) = \frac{N_t(o, a, o')}{N_t(o, a)}$ for $P(o'|o,a)$. We consider two confidence sets for $P$. The first one uses a \emph{time-uniform} variant of Weissman's concentration inequality \citep{weissman2003inequalities} and is defined as follows \citep{maillard2019mathematics,asadi2019model}:
$$
C^{1}_{t,\delta}(o,a) = \Big\{ P'\in \Delta_{\cO}: \|\widehat P_t(\cdot|o,a)  - P' \|_1 \leq \beta_{N_t(o,a)}(\delta) 
\Big\}$$
and 
$
C^{1}_{t,\delta} = \cap_{o,a} C^1_{t,\delta}(o ,a),
$
where for $n\!\in\!\NN$, $\beta_n(\delta): = \sqrt{\tfrac{2}{n}\big(1+\tfrac{1}{n}\big)\log\big(\sqrt{n+1}\tfrac{2^{O}-2}{\delta}\big)}.$ 
By construction, Lemma 
\ref{lem:time-uniform-Laplace-L1} guarantees that uniformly for all $t$, $P\in C^1_{t,2\delta/OA}$, with probability at least $1-\delta/2$, that is, $\PP(\exists t\in \NN: P\notin C^1_{t,\delta/2OA}) \le \delta/2$. 

The second confidence set is based on Bernstein's inequality combined with a peeling technique \citep{maillard2019mathematics}, and is defined as follows:
\als{
C^{2}_{t,\delta}(o,a,o') = \bigg\{ u\in[0,1]: |\widehat P_{t}(o'|o,a)  - u|&\leq \sqrt{\frac{2u(1-u)}{N_t(o,a)}\beta'_{N_t(o,a)}(\delta)} + \frac{\beta'_{N_t(o,a)}(\delta)}{3N_t(o,a)}\bigg\}\,, 
}
and $C^{2}_{t,\delta} = \cap_{o,a,o'} C^2_{t,\delta}(o ,a,o')$, where for $n\!\in\!\NN$ and $\delta\!\in\!(0,1)$, $\beta'_n(\delta):= \eta\log\Big(\frac{\log(n+1)\log(n\eta)}{\delta\log^2(\eta)}\Big)$, where $\eta\!>\!1$ is an arbitrary choice. (We set $\eta\!=\!1.12$, as suggested by \citet{maillard2019mathematics}, to get a small bound.)  Further, let $u_{o,a,o'} :=u_{o,a,o'}(t,\delta)$ be any solution to 
$$
|\widehat P_{t}(o'|o,a)  - u| = \sqrt{\frac{2u(1-u)}{N_t(o,a)}\beta'_{N_t(o,a)}(\delta)} + \frac{\beta'_{N_t(o,a)}(\delta)}{3N_t(o,a)},\qquad u\in [0,1],
$$
which can be found via, e.g., bisection. By construction, $C^{2}_{t,\delta/4O^2A}$ traps $P$ with high probability, uniformly for all $t$: $\PP(\exists t\in \NN: P\notin C^2_{t,\delta/4O^2A})\!\le\!\delta/2$; see 
Lemma \ref{lem:Bernstein_peeling}.

\subsubsection{Confidence Sets for RM State Dynamics $\tau$}
Formally, under a given algorithm and for any $q,q'\in \cQ$ and $\sigma\in \cE_q$, let $N_t(q,\sigma,q')$ denote the number of times a visit to $(q, \sigma)$ was followed by a visit to $q'$, up to time $t$: $N_t(q, \sigma,q') := \sum_{i=1}^{t-1}\mathbb I_{\{(q_{i},\sigma_{i},q_{i+1})=(q,\sigma,q')\}}$. Further, $N_t(q, \sigma) := \max\{1, \sum_{o'}N_t(q,\sigma, q')\}$. Using the observations collected up to $t\ge 1$, we define the empirical estimate $\widehat \tau_t(q'|q,\sigma) = \frac{N_t(q, \sigma, q')}{N_t(q, \sigma)}$ for $\tau(q'|q,\sigma)$.

Similarly to the case of $P$, we consider two confidence sets for $\tau$. The first one is built using  time-uniform Weissman's concentration inequality:
$$
D^{1}_{t,\delta}(q,\sigma) = \Big\{ \tau'\in \Delta_{\cQ}: \|\widehat \tau_t(\cdot|q,\sigma)  - \tau' \|_1 \leq \beta''_{N_t(q,\sigma)}(\delta) 
\Big\}$$
and 
$
D^{1}_{t,\delta} = \cap_{q,\sigma} D^1_{t,\delta}(q ,\sigma),
$
where for $n\!\in\!\NN$, $\beta''_n(\delta): = \sqrt{\tfrac{2}{n}\big(1+\tfrac{1}{n}\big)\log\big(\sqrt{n+1}\tfrac{2^{Q}-2}{\delta}\big)}.$ It follows by construction that  $\PP\Big(\exists t\in \NN: \tau\notin D^1_{t,\delta/2QE}\Big) \le \delta/2$; see 
Lemma \ref{lem:time-uniform-Laplace-L1}. 

The second confidence set is based on Bernstein's inequality (combined with a peeling technique) and is defined as follows: 
\als{
D^{2}_{t,\delta}(q,\sigma,q') = \bigg\{ \lambda\in[0,1]: |\widehat \tau_{t}(q'|q,\sigma)  - \lambda|&\leq \sqrt{\frac{2\lambda(1-\lambda)}{N_t(q,\sigma)}\beta'_{N_t(q,\sigma)}(\delta)} + \frac{\beta'_{N_t(q,\sigma)}(\delta)}{3N_t(q,\sigma)}\bigg\}\,, 
}
and $D^{2}_{t,\delta} = \cap_{q,\sigma,q'} D^2_{t,\delta}(q ,\sigma,q')$. 
Further, let $\lambda_{q,\sigma,q'} :=\lambda_{q,\sigma,q'}(t,\delta)$ be any solution to 
$$
|\widehat \tau_{t}(q'|q,\sigma)  - \lambda|= \sqrt{\frac{2\lambda(1-\lambda)}{N_t(q,\sigma)}\beta'_{N_t(q,\sigma)}(\delta)} + \frac{\beta'_{N_t(q,\sigma)}(\delta)}{3N_t(q,\sigma)}, \qquad \lambda\in [0,1].
$$
Lemma \ref{lem:Bernstein_peeling}  
ensures that 
$\PP\Big(\exists t\in \NN: \tau\notin D^2_{t,\delta/4Q^2E}\Big)\!\le\!\delta/2$.

\subsubsection{Set of Plausible Models}Either choice of confidence sets introduced above, $(C^1,D^1)$ or $(C^2,D^2)$, yields a set of  MDPRMs that are plausible with the collected data up to any time step. More formally, consider a time step $t\ge 1$ and a confidence parameter $\delta\in (0,1)$. Given a set $D$, let $\cR_D$ be the set of RMs defined using transition functions collected in $D$, i.e., $\cR_D:=\{R'=(\cQ,2^{\AtomProp},\tau',\nu): \tau'\in D\}$. Equipped with this, we build the set of MDPRMs 
\begin{align*}
    \cM_{t,\delta} &:= \Big\{M'=(\cO,\cA, P', \cR',{\AtomProp},\eventFunction): P'\in C, \cR'\in \cR_D\Big\}
\, ,
\end{align*}
where $(C,D)=\big(C^1_{t,\delta/2OA},D^1_{t,\delta/2QE}\big)$ or $(C,D)=\big(C^2_{t,\delta/4O^2A},D^2_{t,\delta/4Q^2E}\big)$. Let $\cM^1_{t,\delta}$ and $\cM^2_{t,\delta}$ denote the respective set of MDPRMs. 
This construction ensures that the true MDPRM $M$ belongs to both $\cM^1_{t,\delta}$ and $\cM^2_{t,\delta}$ with high probability, uniformly for all $t$, as formalized in the following lemma:
\begin{lemma}
\label{lem:CI_has_true_MDPRM}
For any $\delta\in (0,1)$: (i) $ \PP(\exists t\in \NN: M\notin \cM^1_{t,\delta}\big) \le \delta$ and (ii) $\PP(\exists t\in \NN: M\notin \cM^2_{t,\delta}\big) \le \delta$.
\end{lemma}
Lemma \ref{lem:CI_has_true_MDPRM} relies on the equivalence between any candidate MDPRM $M'\in \cM_{t,\delta}$ and its associated cross-product MDP $M'^\times=(\cS,\cA,{P'}^\times,\cpR)$, where ${P'}^\times$ and $\cpR$ are defined similarly to (\ref{eq:Mcp_PRM}). Let $\cM^{\times,1}$ and $\cM^{\times,2}$ be the respective set of cross-product MDPs associated to $\cM^1$ and $\cM^2$. 

The special structure of MDPs in $\cM^{\times,1}$ and $\cM^{\times,2}$ may pose some technical challenge since computing an optimal policy over such bounded-parameter MDPs would involve solving bilinear optimizations, which are NP-hard. To accommodate this situation, we introduce proper surrogate sets for $\cM^{\times,1}$ and $\cM^{\times,2}$. 

Let $\overline\nu$ denote the mean of the reward function $\nu(q,\sigma)$ and let us assume that it is known to the agent. (We relax this assumption in Appendix \ref{app:further_algo_detail}.)
Consider the empirical estimate $\widehat P^\times$ defined as follows: For all $s=(q,o)$, $s'=(q',o')$, and $a$, $
\widehat P^\times_t\big(s'|s, a\big) = \widehat P_t(o'|o,a) \widehat \tau_t(q'|q,\sigma),
$
with $\sigma=L(o,a)$. Consider the following sets of MDPs:
\begin{align*}
    {\widetilde \cM}^{\times,1}_{t,\delta} &= \Big\{(\cS,\cA,P'^\times,\cpR): P'^\times \in \cP^1_{t,\delta}\Big\}, \qquad  \hbox{with} \\
  \cP^1_{t,\delta} &:= \bigcap_{s,a}\bigg\{p\in\Delta_{\cS}:\big\|p - \widehat P^\times_t(\cdot|s, a)\big\|_1 \le \beta_{N_t(o,a)}\big(\tfrac{\delta}{2OA}\big) + \beta''_{N_t(q,\sigma)}\big(\tfrac{\delta}{2QE}\big)  \bigg\}    
\end{align*}
Further, 
\begin{align*}
{\widetilde \cM}^{\times,2}_{t,\delta} &= \Big\{(\cS,\cA,P'^\times,\cpR): P'^\times \in \cP^2_{t,\delta}\Big\}, 
\end{align*}
where $\cP^2_{t,\delta} = \bigcap_{s,a,s'} \big\{z\in[0,1]: \big|z  - \widehat P^\times_t(s'|s, a)\big| \le f\big\}$, with 
\begin{align*}
f&:= \widehat \tau_t(q'|q,\sigma)\sqrt{\tfrac{2u_{o,a,o'}(1-u_{o,a,o'})}{N_t(o,a)}\beta'_{N_t(o,a)}\big(\tfrac{\delta}{4O^2A}\big)} + \tfrac{\widehat \tau_t(q'|q,\sigma)}{3N_t(o,a)}\beta'_{N_t(o,a)}\big(\tfrac{\delta}{4O^2A}\big)\\
&\quad + u_{o,a,o'}\sqrt{\tfrac{2\lambda_{q,\sigma,q'}(1-\lambda_{q,\sigma,q'})}{N_t(q,\sigma)}\beta'_{N_t(q,\sigma)}\big(\tfrac{\delta}{4Q^2E}\big)} + \tfrac{u_{o,a,o'}}{3N_t(q,\sigma)}\beta'_{N_t(q,\sigma)}\big(\tfrac{\delta}{4Q^2E}\big)
\end{align*}
We have: 
\begin{lemma}\label{lem:inclusion_model_sets}
For all $\delta\in (0,1)$ and all $t$, $\cM^{\times,1}_{t,\delta} \subseteq {\widetilde \cM}^{\times,1}_{t,\delta}$ and $\cM^{\times,2}_{t,\delta} \subseteq {\widetilde \cM}^{\times,2}_{t,\delta}$.
\end{lemma}
Lemma \ref{lem:inclusion_model_sets} implies that ${\widetilde \cM}^{\times,1}_{t,\delta}$ and ${\widetilde \cM}^{\times,2}_{t,\delta}$ may be used as surrogate for $\cM^{\times,1}_{t,\delta}$ and $\cM^{\times,2}_{t,\delta}$, respectively. In view of Lemma \ref{lem:CI_has_true_MDPRM}, both sets trap the true $\cpM$ with high probability.

\subsection{From Confidence Sets to Algorithms: \UCRLSAL\ and \UCRLSAB}
Equipped with the confidence sets presented above, we are ready to present an algorithm, called \UCRLSA, for learning in MDPRMs. We consider two variants of \UCRLSA\ depending on which confidence set is used: The variant using $(C^1,D^1)$, called \UCRLSAL, can be seen as an extension of \UCRL\ \citep{jaksch2010near} to MDPRMs. Whereas the one built using $(C^2,D^2)$, which we call \UCRLSAB, extends \UCRL-style algorithms with Bernstein's confidence sets (in, e.g., \citep{bourel2020tightening,fruit2020improved,fruit2018efficient}) to MDPRMs. The two algorithms hinge on the same design principle, and hence the same skeleton, but they differ in the choice of the confidence sets as well as their internal procedure of policy computation. As a result, they achieve different regret bounds. In the sequel, we shall use \UCRLSA\ to refer to both variants, but will make specific pointers to each  when necessary. 

\UCRLSA\ implements a form of the \emph{optimism in the face of uncertainty} principle, but in an efficient manner, for MDPRMs. Similarly to many model-based approaches developed based on this principle, it proceeds in internal episodes (indexed by $k\in \NN$) of varying lengths, where within each episode the policy is kept unchanged. Specifically, letting $t_k$ denote the first step of episode $k$, \UCRLSA\ considers a set of plausible MDPRMs and seeks a policy $\pi_{k}:\cS\to\cA$ that has the largest gain over all possible deterministic policies among all such candidate MDPRMs. Practically, as in \UCRL, it suffices to find a $\frac{1}{\sqrt{t_k}}$-optimal solution to the following optimization problem: 
$
\max_{M'\in \cM_{t_k,\delta}, \pi\in \Pi_{M'}} g^{\pi}(M')\, ,
$
where $g^\pi(M')$ denotes the gain of policy $\pi$ in MDPRM $M'$.

In order to solve the optimization problem above, we will be working in the space of cross-product MDPs that correspond to candidate MDPRMs. However, as noted  earlier, rather than considering the sets $\cM^{\times,1}_{t_k,\delta}$ or $\cM^{\times,2}_{t_k,\delta}$, we will consider their surrogate sets $\widetilde \cM^{\times,1}_{t_k,\delta}$ (for \UCRLSAL) or $\widetilde \cM^{\times,2}_{t_k,\delta}$ (for \UCRLSAB). Specifically, to determine the policy $\pi_k$ to be used in episode $k$, we will solve 
$$
\max_{M'\in \cX, \pi\in \Pi_{M'}} g^{\pi}(M')\, , \qquad \hbox{with}\quad \cX\in \Big\{\widetilde \cM^{\times,1}_{t_k,\delta}, \widetilde \cM^{\times,2}_{t_k,\delta}\Big\}\,.
$$
In view of Lemma \ref{lem:inclusion_model_sets}, optimism is guaranteed despite use of $\widetilde \cM^{\times,1}$ and $\widetilde \cM^{\times,2}$ in lieu of $\cM^{\times,1}$ and $\cM^{\times,2}$. This problem can be solved efficiently using  a variant of the \EVI\ algorithm of \citet{jaksch2010near}. More precisely, \EVI\ here takes the following form: 
\begin{align}\label{eq:EVI_upd}
    u^{(i+1)}(s) &= \max_{a\in \cA} \left(\cpR(s,a) + \max_{P'^\times(\cdot|s,a)\in \cP} \sum_{s'}P'^\times(s'|s,a)u^{(i)}(s')\right) \,,\qquad \forall s\in \cQ\times\cO,
\end{align}
starting from an arbitrary choice of $u^{(0)}$. The pseudo-code of \EVI\ is provided in Algorithm \ref{alg:EVI}. The inner maximization in (\ref{eq:EVI_upd}) can be solved exactly, but using different procedures depending on whether $\widetilde \cM^{\times,1}_{t_k,\delta}$ or $\widetilde \cM^{\times,2}_{t_k,\delta}$ is used. Note that we have $\cP=\cP^1_{t,\delta}$ for \UCRLSAL, and $\cP=\cP^2_{t,\delta}$ for \UCRLSAB. We provide these details in Appendix 
\ref{app:further_algo_detail}. 

\EVI\ 
returns a policy $\pi_k$, which is guaranteed to be $\frac{1}{\sqrt{t_k}}$-optimal. \UCRLSA\ commits to $\pi_k$ for $t\!\ge\!t_k$ until the number of observations on some pair $(o,a)$ or $(q,\sigma)$ is doubled.\footnote{This is inspired by the stopping criterion in UCRL2 \citep{jaksch2010near}.} More precisely, the sequence $(t_k)_{k\ge 1}$ satisfies: $t_1\!=\!1$, and for $k\ge 1$, 
\begin{align*}
t_{k}\!=\!\min\left\{t>t_{k-1}:\max_{o,a}\frac{\sum_{t'=t_{k-1}}^{t}\mathbb I_{\{(o_{t'},a_{t'})=(o,a)\}}}{N_{t_{k-1}}(o,a)}\ge 1\,\, \hbox{ or } \,\, \max_{q,\sigma\in \cE_q}\frac{\sum_{t'=t_{k-1}}^{t}\mathbb I_{\{(q_{t'},\sigma_{t'})=(q,\sigma)\}}}{N_{t_{k-1}}(q,\sigma)}\ge 1\right\}.
\end{align*}
The pseudo-code of \UCRLSA\ is presented in Algorithm \ref{alg:ucrlSA}. We recover \UCRLSAL\ (resp.~\UCRLSAB) if $\widetilde \cM^{\times,1}_{t,\delta}$ (resp.~$\widetilde \cM^{\times,2}_{t,\delta}$) is used. 



\begin{algorithm}[!hbtp]
   \caption{\UCRLSA}
   \label{alg:ucrlSA}
   \footnotesize
\begin{algorithmic}
    \REQUIRE $\cO,\cA,\delta$
   \STATE \textbf{Initialize:} For all $(o,a,o')$, set $N_0(o,a)=0$, $N_0(o, a, o')=0$ and $\counter_0(o,a)=0$. For all $q,q'\in \cQ$ and $\sigma\in \cE_q$, set $N_0(q,\sigma)=0$, $N_0(q, \sigma, q')=0$ and $\counter_0(q,\sigma)=0$. Set $t_0=0$, $t=1$, $k=1$, and observe the initial state $s_1 = (q_1, o_1)$
   \FOR{episodes $k\geq  1$}
       \STATE Set $t_k = t$
       \STATE Set $N_{t_k}(o,a) = N_{t_{k-1}}(o,a)+ \counter_{k}(o,a)$ for all $(o,a)$ and $N_{t_k}(q,\sigma) = N_{t_{k-1}}(q,\sigma)+ \counter_{k}(q,\sigma)$ all $q\in \cQ$ and $\sigma\in \cE_q$
       \STATE Set $\counter_k(o,a) = 0$ for all $(o, a)$ and $\counter_k(q,\sigma) = 0$ for all $q\in \cQ$ and $\sigma\in \cE_q$
       \STATE Compute empirical estimates $\widehat P_{t_k}(\cdot|o,a)$ for all $(o,a)$ and $\widehat \tau_{t}(\cdot|q, \sigma)$ for all $q\in \cQ$ and $\sigma\in \cE_q$
       \STATE Compute  $\pi_k = \texttt{EVI}\big(\cP, \tfrac{1}{\sqrt{t_k}}\big)$. 
       \STATE \qquad \emph{(Set $\cP=\cP^1_{t,\delta}$ for \UCRLSAL, and $\cP=\cP^2_{t,\delta}$ for \UCRLSAB.)} 
       \WHILE{$\counter_{k}(o_t,\pi_{k}(q_t, o_t))<  \max\{1,N_{t_k}(o_t,\pi_k(q_t, o_t))\}$ and $\counter_{k}(q_t,\eventFunction(o_t, \pi_{k}(q_t, o_t)))<  \max\{1,N_{t_k}(q_t,L(o_t, \pi_{k}(q_t, o_t)))\}$}
            \STATE Play action $a_t=\pi_{k}(q_t,o_t)$
            \STATE Collect label $\sigma_t=L(o_t,a_t)$            
            \STATE Receive next observation $o_{t+1}\sim P(\cdot|o_t, a_t)$ and next state $q_{t+1} \sim \tau(\cdot|q_t, \sigma_t)$
            \STATE Receive reward $r_t\sim \nu(q_t,\sigma_t)$
            \STATE Set $N_{t+1}(o_t,a_t,o_{t+1})=N_t(o_t,a_t,o_{t+1})+1$ and $N_{t+1}(q_t,\sigma_t,q_{t+1})=N_t(q_t,\sigma_t, q_{t+1})+1$
            \STATE Set $\counter_k(o_t,a_t)=\counter_k(o_t,a_t)+1$ and $\counter_k(q_t,\sigma_t)=\counter_k(q_t,\sigma_t)+1$
            \STATE Set $t=t+1$
       \ENDWHILE
   \ENDFOR
\end{algorithmic}
\normalsize
\end{algorithm}


\begin{algorithm}[!hbtp]
	\caption{\texttt{EVI}$(\cP,\epsilon)$}
\small
	\label{alg:EVI}
	\begin{algorithmic}
		\STATE \textbf{Initialize:} $u^{(0)}\equiv 0, u^{(-1)}\equiv-\infty$, $i=0$ 
		\WHILE{$\max_{s\in \cS}(u^{(i)}-u^{(i-1)})(s) - \min_{s\in \cS}(u^{(i)}-u^{(i-1)})(s) > \epsilon$}
        \STATE Get $\widetilde{P}^\times$ using \texttt{MAXP-L1} (Algorithm S1, for \UCRLSAL) or \texttt{MAXP-B} (Algorithm S2, for \UCRLSAB)
         

        \STATE For all $s \in \cQ \times \cO$, update: $$u^{(i+1)}(s) = \max_{a\in \cA} \left(\cpR(s,a) + \sum_{s'}\widetilde{P}^\times(s'|s,a)u^{(i)}(s')\right)$$


		\STATE Set $i=i+1$
\ENDWHILE
\STATE \textbf{Output:} 
$$
\pi(s) =  \arg\max_{a\in \cA} \left(\cpR(s,a) + \sum_{s'}\widetilde{P}^\times(s'|s,a)u^{(i)}(s')\right),\qquad \forall s \in \cQ \times \cO$$
	\end{algorithmic}
\normalsize
\end{algorithm}



\section{Theoretical Regret Guarantees}\label{sec:regret_bound}
In this section, we present finite-time regret bounds for the two variants of \UCRLSA\ that hold with high probability. We present regret bounds for both probabilistic and deterministic RMs. First, we introduce a notion of diameter that renders relevant for RMs. 

\subsection{RM-Restricted Diameter}
As in tabular MDPs, regret performance of an RL algorithm in average-reward MDPRMs would depend on some measure of connectivity. Specifically, for MDPRMs with communicating cross-product MDPs, diameter notions render most relevant. We distinguish between two notions of diameter for MDPRMs. The first one, denoted by $\Dcp$, is defined as the diameter of the cross-product MDP associated to the considered MDPRM, coinciding with the classical definition of diameter, which we recall below for completeness: 
\begin{definition}[\cite{jaksch2010near}]
    The diameter $\Dcp$ of an MDP $\cpM$ is defined as 
    $$
    \Dcp=\max_{s\neq s'}\min_\pi \EE[T^\pi(s,s')],$$ 
    where $T^\pi(s,s')$ is the number of steps it takes to reach $s'\in \cS$ starting from $s\in \cS$ and following policy $\pi:\cS\to\cA$.  
\end{definition}

The second one is a novel notion, which we shall call \emph{RM-restricted diameter}, and is tailored to the structure of MDPRMs. To formalize it, let us introduce for $s=(q,o)\in \cS$, 
$$
\cB_{s} := \bigcup_{a} \Big\{q'\in \cQ: \tau\big(q'\big|q,  \eventFunction(o, a)\big)>0\Big\}.
$$
Intuitively, for a given $s=(q,o)$, $\cB_s\subseteq \cQ$ collects all possible next-states of the RM that can be reached from $q$ via the \emph{detectable labels} in $o$. In the worst-case, one has $\cB_s = \cQ$ for some state $s$. However, many high-level tasks in practice often admit RMs with sparse structures, where $\cB_s$ may be a small subset of $\cQ$ (cardinality-wise). Equipped with this, we define the \emph{RM-restricted diameter} for state $s=(q,o)\in \cS$:  

\begin{definition}[RM-Restricted Diameter]
\label{def:diameter_SAMDP_1}
	Consider state $s\!=\!(q, o)\!\in\!\cS$. For $s_1,s_2\in \cB_{s}\!\times\!\cO$ with $s_1\neq s_2$, let $T^\pi(s_1,s_2)$ denote the number of steps it takes to reach $s_2$ starting from $s_1$ and following policy $\pi:\cS\to\cA$. The \emph{RM-restricted diameter} of MDPRM $M$ for state $s$ is defined as
	$$
	D_{s}:= \max_{s_1,s_2\in \cB_{s} \times \cO}   \min_\pi \Esp[T^\pi(s_1,s_2)].
	$$
\end{definition}

It is evident that $D_{s}\!\le\!\Dcp$ for all $s\!\in\!\cS$, in view of  $\cB_s\!\subseteq\!\cQ$. Further, if $\cB_s=\cQ$ for some state $s$, then the RM-restricted diameter for $s$ coincides with $\Dcp$. Since $\cB_s$ might be a proper (and possibly small) subset of $\cQ$, $D_s$ is a problem-dependent refinement of $\Dcp$. It is worth noting that a small $\cB_s$ does not necessarily imply that $D_s\!\ll\!\Dcp$ as $D_s$ is determined by both $\cB_s$ and the transition function $\cpP$ of $\cpM$. Interestingly, however, there exist cases where $D_{s}\lesssim \Dcp/Q$, as we illustrate next.

\begin{figure}[t]
  \label{fig:illustrative_diameter_1}
\centering
\scriptsize
\def\svgwidth{.85\linewidth}
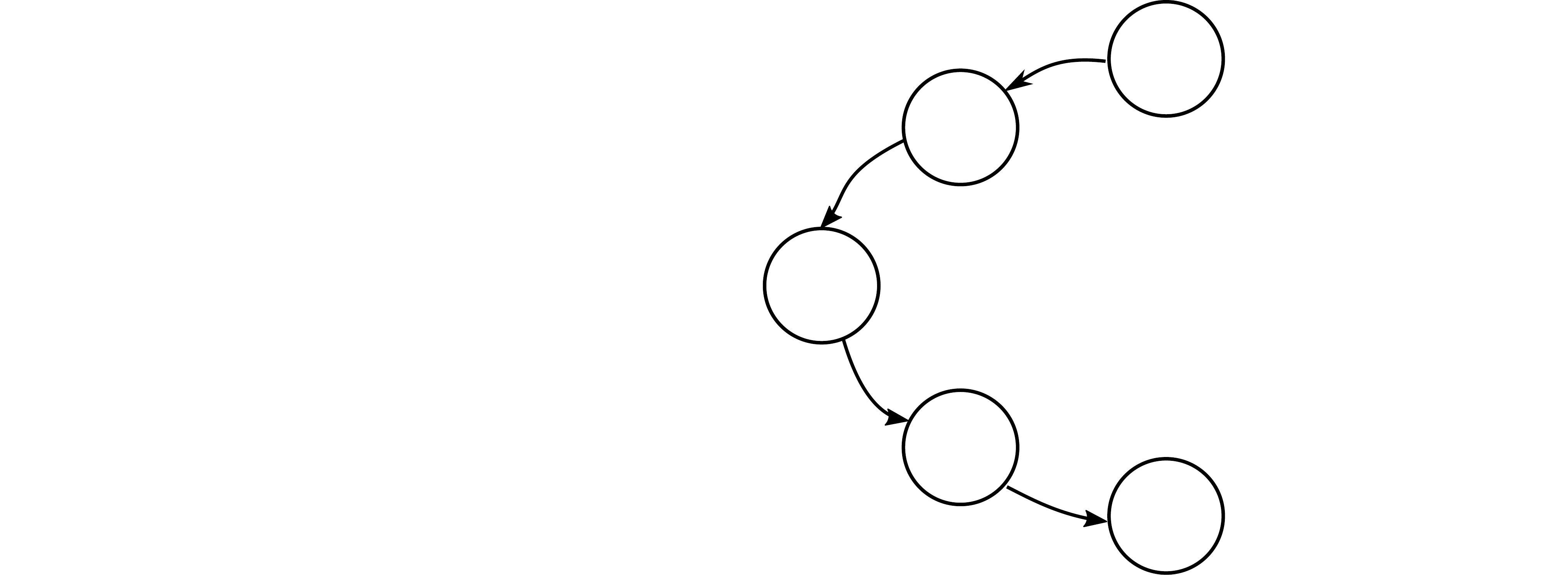
  {\caption{An example where RM-restricted diameter $D_s \lesssim \Dcp/Q$. The labeled MDP in left, and the RM in right.}
  \label{fig:restricted_diameter}}
\end{figure}

Consider the MDPRM shown in Figure \ref{fig:restricted_diameter}, where there are two observation states $o_0$ and $o_1$, with identical transition probabilities parameterized by $\delta\in (0,\tfrac{1}{2})$. In $o_0$, there is one action, but no event. In $o_1$, there are two actions: $a_0$ resulting in detecting $\sigma_A$, and $a_1$ that leads to $\sigma_B$. The underlying (deterministic) RM has $Q$ states arranged in a cycle, such that $\sigma_A$ and $\sigma_B$ yield transitions in the clockwise and counter-clockwise directions, respectively. It holds that 
for all $q\in \cQ$, $D_{o_1,q} = \tfrac{2}{\delta} + 1 + \tfrac{\delta}{1 - \delta}$ and $D_{o_0,q}= \tfrac{1}{\delta}$, whereas $\Dcp = \frac{\lfloor Q/2\rfloor}{\delta} + 1 + \frac{\delta}{1 - \delta}$. 
Thus, while $\Dcp$ grows as $\tfrac{Q}{\delta}$, $D_s$ for all $s\!\in\!\cS$ will be $\tfrac{1}{\delta}$. In summary, $D_s \lesssim \Dcp/Q$.  

\subsection{Regret Bounds}

\subsubsection{Probabilistic Reward Machines}
First, we present regret bounds for \UCRLSA\ in the case of MDPRMs with probabilistic RMs. 

The following theorem provides a regret bound for \UCRLSAL, which is constructed using $(C^1,D^1)$: 
\begin{theorem}
\label{thm:regret_UCRLSAL1}
		Under \UCRLSAL, with probability higher than $1-3\delta$ and uniformly over all $T\ge 2$,
		\als{
            \kR(T) \lesssim 	\Dcp\sqrt{OAT\big(O+\log(T/\delta)\big)}+ \Dcp\sqrt{\sum_{q\in \cQ} E_q T\big(Q+\log(T/\delta)\big)} + \Dcp (OA + QE)\log(T)
            }
\end{theorem}

To present a regret bound for \UCRLSAB\ (constructed using $(C^2,D^2)$), we introduce some necessary notations. For $(o,a)\in \cO\times \cA$, we let  $K_{o,a}$ be the number of possible next observations  in $\cO$ under $(o,a)$, that is, $K_{o,a}:=|\{o'\in \cO:P(o'|o,a)>0\}|$. Similarly, for $q\in \cQ$ and $\sigma\in \cE_q$, we let  $K_{q,\sigma}$ be the number of possible next RM-states in $\cQ$ under $(q,\sigma)$, that is, $K_{q,\sigma}:=|\{q'\in \cQ:\tau(q'|q,\sigma)>0\}|$. 

\begin{theorem}
\label{thm:regret_UCRLSAB}
		Under \UCRLSAB, uniformly over all $T\ge 2$, with probability higher than $1-3\delta$,
				\als{
			\kR(T) &\lesssim  \Dcp \sqrt{\sum_{o,a}K_{o,a}T\log(\log(T)/\delta)} + \Dcp \sqrt{\sum_{q,\sigma\in \cE_q}K_{q,\sigma}T\log(\log(T)/\delta)}\\
            &+ \Dcp Q^2O^2AE \log(T)\log(\log(T)) \, . 
		}			
\end{theorem}

\subsubsection{Deterministic Reward Machines} Now we restrict attention to the special class of deterministic RMs and present improved regret bounds. In the case of deterministic RMs, it is evident that there will be no need to maintain a confidence set for $\tau$. Further, it will  no longer be necessary to use a surrogate set of models. Hence, the variants of \UCRLSA\ reduce to their respective form of \texttt{UCRL-RM}, which were initially presented in our previous work  \citep{bourel2023exploration}. We will refer to this special case of \UCRLSA\ as \texttt{UCRL-RM}, to comply with the terminology used in \citep{bourel2023exploration}. 


\begin{theorem}
\label{thm:regret_UCRLSAL1_deterministic}
		Given an MDPRM $M$, let $\mathbf c_M =\sum\limits_{o\in \cO} \max\limits_{q\in \cQ}D_{q,o}^2$. Uniformly over all $T\ge 2$, with probability higher than $1-3\delta$, the regret under \texttt{UCRL-RM-L1} in $M$ satisfies
				\als{
			\kR(T) \lesssim \sqrt{\mathbf c_M AT\big(O + \log(T/\delta)\big)} + \Dcp \sqrt{T\log(T/\delta)}\, .
		}
\end{theorem}

\begin{theorem}
\label{thm:regret_UCRLSAB_deterministic}
		Given an MDPRM $M$, let $\mathbf c'_M =\sum\limits_{o\in \cO,a\in \cA} K_{o,a}\max\limits_{q\in \cQ}D_{q,o}^2$. Uniformly over all $T\ge 2$, with probability higher than $1-3\delta$, the regret under \texttt{UCRL-RM-B} in $M$ satisfies
				\als{
						\kR(T) \lesssim \sqrt{\mathbf c'_M T\log(\log(T)/\delta)} + \Dcp \sqrt{T\log(\log(T)/\delta)} \,. 
		}	
\end{theorem}

Here, $\mathbf c_M$ and ${\mathbf c}'_M$ are problem-dependent quantities that reflect the contribution of various states to the regret, weighted by their associated RM-restricted diameter. In the worst-case, $\mathbf c_M \le O\Dcpsq$ and $\mathbf c'_M = \Dcpsq\sum_{o,a}K_{o,a}$. But in view of the example earlier, there are problem instances in which $\mathbf c_M\lesssim O\Dcpsq/Q^2$ and  $\mathbf c'_M\lesssim \Dcpsq/Q^2\sum_{o,a}K_{o,a}$. Therefore, these quantities could adapt to the sparsity structure of the underlying MDPRM. In contrast, the reported regret bounds for the case of probabilistic RMs would scale with the diameter $\Dcp$ of the cross-product MDP, which is structure-oblivious and (potentially much) larger. It is an interesting direction to derive similar regret bounds for probabilistic RMs.



\subsection{Discussion} 
Any algorithm available for tabular RL could be directly applied to $\cpM$, obliviously to the structure induced by the RM. In particular, \UCRL\ \citep{jaksch2010near} attains a regret of scaling as $\Dcp OQ\sqrt{AT \log{T}}$, although its regret with {improved confidence sets used here} would grow as $\Dcp\sqrt{AOQT(OQ + \log{T})}$. Further, \UCRLBernstein\ achieves a regret of $O(\Dcp\sqrt{T\log(\log(T)) \sum_{o,a}K_{q,o,a}}))$.\footnote{A factor of $\sqrt{\Dcp/\log(T)}$ can be shaved off the regret of UCRL2-B as reported by \citet{fruit2020improved}, and the same improvement may carry over to \UCRLSAB. We exclude comparisons to EBF introduced by \citet{zhang2019regret} growing as $O(\sqrt{\Dcp QOAT\log(T)})$, as it does not admit an efficient implementation.}

It would render natural to compare \UCRLSAL\ with \UCRL, and \UCRLSAB\ with \UCRLBernstein, in terms of regret dependency on problem parameters $O,Q$, and $A$. Under \UCRLSAL, the regret order is $\sqrt{(OA+\sum_q E_q)T\log(T)}$ for large time horizon $T$ (relative to $O$ and $Q$), whereas it is or $\sqrt{(O^2A+Q\sum_qE_q)T}$ for moderate $T$. In contrast, for \UCRL\ it scales as $OQ\sqrt{AT\log(T)}$. It is clear that a dependency on $QO$ is improved to one of the form $Q+O$, especially when $E_q\ll Q$ and $A\ll O$, which reasonably hold in practical situations. A similar remark holds when comparing to the improved regret of \UCRL. In the case of \UCRLSAB\ compared with \UCRLBernstein, a dependency on the (cumulative) support size of $\cpP$ is traded with $K_{o,a} + K_{q,\sigma}$, that is, the support sizes of $P$ and $\tau$. The quantities $K_{o,a}$ and $K_{q,\sigma}$ are more capable of representing the sparsity of MDPRM than the support size of $\cpP$. 

In the case of deterministic RMs, \texttt{UCRL-RM-L1} improves over \UCRL\ by a multiplicative factor of $Q$. However, in some specific instances, we have $D_s\lesssim \Dcp/Q$ for all $s$, so that $\mathbf c_M\sim  O(\Dcp/Q)^2$ for such $M$. On such MDPRMs, for moderate $T$, we obtain an improvement in the regret bound by a multiplicative factor of at least $Q$, but in some examples this can be as large as $Q^2$. For large horizons (relative to $O$), the respective gains over \UCRL\ are $\sqrt{Q}$ and $Q^{3/2}$.  



In view of $D_s\leq \Dcp$, $\mathbf c'_M\le \Dcpsq\sum\nolimits_{o,a}K_{o,a}$. Hence, the regret of \texttt{UCRL-RM-B}, in the worst case grows as $\Dcp\sqrt{\sum\nolimits_{o,a} K_{o,a} T}$. However, in some specific instances, we have $D_s\lesssim \Dcp/Q$ for all $s$, thus yielding $\mathbf c'_M\lesssim  (\Dcp/Q)^2\sum\nolimits_{o,a}K_{o,a}$. On such MDPRMs, its regret is of order 
$
\tfrac{\Dcp}{Q}\sqrt{\sum\nolimits_{o,a} K_{o,a} T}. 
$ 
In summary, \texttt{UCRL-RM-B} improves \UCRLBernstein\ in regret by a factor of, at least, $\sqrt{Q}$. 
Moreover, in instances where $D_s\lesssim \Dcp/Q$, the improvement  could be as large as a factor of $Q^{3/2}$.

\section{Regret Lower Bound}\label{sec:regret_LB}

In this section, we present a regret lower bound for the class of MDPRMs under the communicating assumption on the associated cross-product MDP. For communicating tabular MDPs with $S$ states, $A$ actions, and diameter $D$, a regret lower bound of $\Omega(\sqrt{DSAT})$ is presented by \citet{jaksch2010near}, which relies on a carefully constructed family of worst-case MDPs.  
\emph{However, this does not translate to a lower bound of $\Omega(\sqrt{\Dcp QOAT})$ for the cross-product} $\cpM$ {associated to a given MDPRM $M$.} This is due to the fact that the transition function of the aforementioned worst-case MDPs does not satisfy (\ref{eq:Mcp_PRM}). In other words, \emph{there exist no MDPRMs} for which those worst-case MDPs become their associated cross-product MDPs. In the following theorem, we present a regret lower bound that holds for any MDPRM $M$ with a communicating cross-product $\cpM$. 
\begin{theorem}\label{thm:LB}
For any $O\!\ge\!3$, $A\!\ge\!2$, $Q\!\ge\!2$, and $\Dcp\!\ge\!Q(6+2\log_A(O))$, $T\!\ge\!\Dcp OA$ and $E\!\ge\!4$, there exists a family of MDPRMs with deterministic RMs comprising $O$ observations states, $A$ actions, $Q$ RM states, and diameter $\Dcp$ of the associated $\cpM$, in which the regret of any algorithm $\mathbb A$ satisfies 
$$
\EE[\kR(\mathbb A,T)] \ge c_0\sqrt{\Dcp OAT}, 
$$
where $c_0>0$ is a universal constant. 
\end{theorem}

This theorem asserts  a \emph{worst-case} regret lower bound growing as $\Omega(\sqrt{\Dcp OAT})$ and is provided in 
Appendix \ref{app:regret_LB}. 
To establish this result, we carefully construct an instance of MDPRM. 
In order to make it a worst-case instance, both $P$ and $\cR$ have to be chosen in a way to challenge exploration. To this end, we construct an RM with a non-trivial structure, whereas for $P$, we take inspiration from the worst-case MDPs presented by \citet{jaksch2010near}, so that on the resulting MDPRM, the regret of any algorithm 
grows  as $\Omega(\sqrt{\Dcp OAT})$ even when the RM and associated rewards are known to the learner. 

We remark that in the construction of the worst-case MDPRM, we use a deterministic RM $\cR$. Nevertheless, the lower bound is likely loose for the case of probabilistic RMs. Deriving a worst-case MDPRM with a probabilistic RM is substantially more challenging and is an interesting direction for future research. We finally remark that the lower bound does not contradict our regret bounds; in particular, in the case of deterministic machines, the worst-case instances considered in Theorem \ref{thm:LB} satisfy $\max_q D_{q,o} \simeq \Dcp$.

\section{Conclusion}\label{sec:conclusion}
We studied reinforcement learning in average-reward Markov decision processes with probabilistic reward machines (MDPRMs), in the regret minimization setting. We assumed that the reward machine is unknown but its state is observable. We introduced two algorithms tailored to leverage the structure of MDPRMs, and analyzed their regret. 
Both algorithms significantly outperform existing baselines in terms of theoretical regret guarantees. We also derived a  regret lower bound for MDPRMs with deterministic machines that relies on a novel construction of worst-case MDPRMs. An interesting future work direction is to devise efficient algorithms for MDPRMs when the state of the RM is not observed. Another interesting direction is to improve our lower bound for the class of probabilistic machines. 

\section*{Acknowledgment}Hippolyte Bourel is supported by the Independent Research Fund Denmark, grant number 1026-00397B. Anders Jonsson is partially supported by the Spanish grant PID2019-108141GB-I00 and the European project TAILOR (H2020, GA 952215). Odalric-Ambrym Maillard is supported by the French Ministry of Higher Education and Research, Inria, Scool, the Hauts-de-France region, the MEL and the I-Site ULNE regarding project R-PILOTE-19-004-APPRENF. Mohammad Sadegh Talebi is partially supported by the Independent Research Fund Denmark, grant number 1026-00397B.

\bibliographystyle{unsrtnat}
\bibliography{RM-bib.bib}

\appendix
\section{Proof of Lemma \ref{lem:cross_product}} %
\label{app:some_lemmas}

\bp 
Let $M\!=\!(\cO,\cA,P,\cR,{\AtomProp},L)$ and $\cS=\cQ\times \cO$. 
For any $t\in \NN$, let $h_t:=(s_1,a_1,\ldots,s_{t-1},a_{t-1},s_t)$, where $s_{t'}:=(q_{t'},o_{t'})$. We show that for any $h\in (\cS\times\cA)^{t-1}\times \cS$, $s'=(q',o')\in \cS$, $a\in \cA$, and $B\subseteq [0,1]$:
\begin{align*}
\PP(s_{t+1}=s'|h_t=h,a_t=a)= P(o'|o, a)\tau(q'|q, \eventFunction(o, a))\,,   \quad \PP(r_{t}\in B|h_t=h,a_t=a)&= \nu\big(q, \eventFunction(o, a)\big)(B)\,,    
\end{align*}
thus implying that the state and reward dynamics are fully determined by $(s_t,a_t)$. 
For any $(q',o')\in \cS$, we have
\begin{align*}
    \PP\big(s_{t+1}=(q',o')\big|h_t=h,a_t=a\big)     &= \PP\big(o_{t + 1} = o'\big| h_t=h, a_t = a\big)\PP\big(q_{t+1} = q' \big| h_t=h, o_{t+1} = o', a_t = a\big)  \\
    &= \PP\big(o_{t + 1} = o'\big| o_t = o, a_t = a\big)\PP\big(q_{t+1} = q' \big| s_t = (q,o), o_{t+1} = o', a_t = a\big)  \\
  &= P(o'|o, a)\tau\big(q'|q, \eventFunction(o, a)\big)\, ,
\end{align*}
where the second line follows from the fact that observation dynamics are Markovian and from the definition of RMs.

Moreover, for any set $B\subseteq [0,1]$, we have
\begin{align*}
    \PP\big(r_{t}\in B\big|h_t=h,a_t=a\big) 
       &= \PP\big(r_t \in B\big| s_t=(q,o), o_{t+1}=o',a_t = a\big)  = \nu\big(q, \eventFunction(o,a)\big)(B)\,, 
\end{align*}
thus verifying the two claims, and the lemma follows.
\ep

\section{Properties of the Set of Models}

\subsection{Proof of Lemma \ref{lem:inclusion_model_sets}}

\bp
Consider $\cM^1_{t,\delta}$. Its corresponding set of cross-product MDPs, $\cM^{\times,1}_{t,\delta}$, collects all MDPs $M'=(\cS,\cA,P'^\times,\cpR)$ such that for all $s,s'\in \cS$ and $a\in \cA$, 
$
P'^\times(s'|s,a) = P'(o'|o,a)\tau'(q'|q,\sigma)
$ such that $P'\in C^1_{t,\delta}$ and $\tau'\in D^1_{t,\delta}$. Now, if $M'\in \cM^{\times,1}_{t,\delta}$, then it holds for any $(s,a)$ that
\begin{align*}
\big\|P'^\times(\cdot|s,a)-\widehat P^\times_t(\cdot|s, a)\big\|_1 &\le \big\|P'(\cdot|o,a)-\widehat P_t(\cdot|o, a)\big\|_1 + \big\|\tau'(\cdot|q,\sigma)-\widehat \tau_t(\cdot|q, \sigma)\big\|_1 \\
&\le \beta_{N_t(o,a)}\big(\tfrac{\delta}{2OA}\big) + \beta''_{N_t(q,\sigma)}\big(\tfrac{\delta}{2QE}\big)\,,
\end{align*}
where the first inequality follows from the identity that for probability vectors $a,b,c$, and $d$, $\sum_{i} |a_ib_i - c_id_i|\le \sum_{i}|a_i-c_i| + \sum_i |b_i-d_i|$, and where the last inequality follows from the fact that $P'\in C^1_{t,\delta}$ and $\tau'\in D^1_{t,\delta}$. Thus, $\cM^{\times,1}_{t,\delta} \subseteq {\widetilde \cM}^{\times,1}_{t,\delta}$. 

The second statement can be proven using a very similar argument; its proof is thus omitted.
\ep

\subsection{Proof of Lemma \ref{lem:CI_has_true_MDPRM}}

\bp Recall that ${\cM}^{\times,1}_{t,\delta}$ (resp.~${\cM}^{\times,2}_{t,\delta}$) denote the set of cross-products MDPs associated to MDPRMs in $\cM^1_{t,\delta}$ (resp.~$\cM^2_{t,\delta}$). To prove the lemma, we show that $\cpM$ belongs to the corresponding induced set of cross-product MDPs with high probability. 

\underline{Part (i).}  Note that $\cM^{\times,1}_{t,\delta}$ collects all MDPs $M'=(\cS,\cA,P'^\times,\cpR)$ such that for all $s,s'\in \cS$ and $a\in \cA$, 
$
P'^\times(s'|s,a) = P'(o'|o,a)\tau'(q'|q,\sigma)
$ 
such that $P'\in C^1_{t,\delta}$ and $\tau'\in D^1_{t,\delta}$. It is evident that $M\notin\cM^1_{t,\delta}$ if and only if $\cpM\notin\cM^{\times,1}_{t,\delta}$. Hence, 
\als{
\Pr\big(\exists t\in\Nat, \, M\notin \cM^1_{t,\delta}\big) 
&= \Pr\Big(\exists t\in\Nat, \, \cpM\notin \cM^{\times,1}_{t,\delta}\Big) \\
&\le \Pr\Big(\exists t\in\Nat, \exists P \notin C^1_{t,\delta/2OA} \quad\hbox{or}\quad\exists \tau \notin D^1_{t,\delta/2QE}\Big) \\
&= 
\Pr\Big(\exists t\in\Nat, \exists(o,a)\in\cO\times\cA, P(\cdot|o,a) \notin C^1_{t,\delta/2OA}(o,a)\Big) \\
&+ \Pr\Big(\exists t\in\Nat, \exists q\in \cQ,\sigma\in \cE_q, \tau(\cdot|q,\sigma)\notin D^1_{t,\delta/2QE}(q,\sigma)\Big) \\
&\le \sum_{o\in \cO,a\in \cA} \Pr\Big(\exists t\in\Nat, P(\cdot|o,a) \notin C^1_{t,\delta/2OA}(o,a)\Big) \\
&+ \sum_{q\in \cQ,\sigma\in \cE} \Pr\Big(\exists t\in\Nat, \tau(\cdot|q,\sigma) \notin D^1_{t,\delta/2QE}(q,\sigma)\Big)\\
&\le \delta\, ,
}
where the last inequality follows from Lemma \ref{lem:time-uniform-Laplace-L1}. 

\underline{Part (ii).} The proof follows similar lines as in that of Part (i). We have:
\als{
\Pr\big(\exists t\in\Nat, \, M\notin \cM^2_{t,\delta}\big) 
&= \Pr\Big(\exists t\in\Nat, \, \cpM\notin \cM^{\times,2}_{t,\delta}\Big) \\
&\le \Pr\Big(\exists t\in\Nat, \exists P \notin C^2_{t,\delta/4O^2A} \quad\hbox{or}\quad \exists \tau \notin D^1_{t,\delta/4Q^2E}\Big) \\
&= 
\Pr\Big(\exists t\in\Nat, \exists(o,a,o')\in\cO\times\cA\times\cO, P(o'|o,a) \notin C^2_{t,\delta/4O^2A}(o,a,o')\Big) \\
&+ \Pr\Big(\exists t\in\Nat, \exists q,q'\in\cQ, \sigma\in \cE_q, \tau(q'|q,\sigma)\notin D^1_{t,\delta/4Q^2E}(q,\sigma,q')\Big) \\
&\le \sum_{o,o'\in \cO,a\in \cA} \Pr\Big(\exists t\in\Nat, P(o'|o,a) \notin C^2_{t,\delta/4O^2A}(o,a,o')\Big) \\
&+ \sum_{q,q'\in \cQ,\sigma\in \cE_q} \Pr\Big(\exists t\in\Nat, \tau(q'|q,\sigma) \notin D^1_{t,\delta/4Q^2E}(q,\sigma,q')\Big)\\
&\le \delta\, ,
}
where the last inequality uses Lemma \ref{lem:Bernstein_peeling}.
\ep

\section{Further Algorithmic Details}\label{app:further_algo_detail}

\subsection{Extended Value Iteration for MDPRMs}
\texttt{EVI} relies on solving the following maximization problem in each round, and for any  $s \in \cQ \times \cO$:

\begin{align}
 \max_{{P'}^\times(\cdot|s,a) \in \cP} \sum_{s'}{P'}^\times(s'|s,a)u(s')
 \label{eq:im-t}
\end{align}
where $u$ is the value function at the current iteration of \EVI, and  where $\cP=\cP^1_{t,\delta}$ for \UCRLSAL, and $\cP=\cP^2_{t,\delta}$ for \UCRLSAB. 
Algorithm \ref{alg:MAXP_L1} finds a solution to problem  (\ref{eq:im-t}) for \UCRLSAL, whereas  Algorithm \ref{alg:MAXP_B} does so for \UCRLSAB. 
Algorithm \ref{alg:MAXP_L1} is quite similar to the one used in UCRL2 \citep{jaksch2010near}, whereas Algorithm \ref{alg:MAXP_B} is used in UCRL2B and similar (e.g., in \citep{dann2015sample}).

\begin{algorithm}[!hbtp]
	\caption{\texttt{MAXP-L1}}
\small
	\label{alg:MAXP_L1}
	\begin{algorithmic}
	\STATE For all $s'\in \cQ \times \cO$, set $p(s') = \widehat{P}^\times(s'|s, a)$
	\STATE $s_{\max} = \arg\max_{s'\in \cQ \times \cO} u(s')$
	\STATE $p(s_{\max}) = \max\Big\{1, \,p(s_{\max}) + \tfrac{1}{2} \left(\beta_{N_t(o,a)}\big(\tfrac{\delta}{2OA}\big) + \beta''_{N_t(q,\sigma)}\big(\tfrac{\delta}{2Q\Sigma}\big)\right)\Big\}$
	\STATE $\cL = \argsort_{s'} u(s')$
    \STATE $\ell = 0$
	\WHILE{$\sum_{o'\in \cO}p(s') > 1$}
	    \STATE $p(\cL_\ell) = \max\Big\{0, \,p(\cL_\ell) + 1 - \sum_{s' \in \cQ \times \cO} p(s')\Big\}$
	    \STATE Set $\ell = \ell + 1$
	\ENDWHILE
    \STATE \textbf{Output:} $\widetilde{P}^\times(\cdot|s, a)=p$
	\end{algorithmic}
\normalsize
\end{algorithm}

\begin{algorithm}[!hbtp]
	\caption{\texttt{MAXP-B}}
\small
	\label{alg:MAXP_B}
	\begin{algorithmic}
	\STATE For all $s'\in \cS \times \cO$, set $p(s') = \min\big\{P'\in C^2_{t,\delta}\big\}$
	\STATE $\cL = \argsort_{s'} u(s')$
    \STATE $\ell = QO-1$
	\WHILE{$\sum_{s'\in \cQ \times \cO}p(s') < 1$}
	    \STATE Set 
     $$p(\cL_\ell) = \min\bigg\{\max\Big\{z\in \widetilde D_{t,\delta}\big(s, a, \cL_\ell\big)\Big\}, 1 - \sum_{s' \in \cS \times \cO} p(s')\bigg\}$$
	    \STATE $\ell = \ell - 1$
	\ENDWHILE
    \STATE \textbf{Output:} $\widetilde{P}^\times(\cdot|s, a)=p$
	\end{algorithmic}
\normalsize
\end{algorithm}

\subsection{Unknown Mean Rewards}
\label{app:unknown_rewards}
Now we discuss the case of unknown mean rewards, i.e., when the agent has no prior knowledge about $\overline \nu$. To accommodate this situation, the agent maintains  confidence sets for the various mean rewards as follows. For $q\in\cQ$ and $\sigma\in\cE_q$, define
\als{
C^{\text{reward}}_{t,\delta}(q,\sigma) &= \Big\{ \lambda\in [0,1]: \big|\widehat \nu_{t}(q,\sigma)  - \lambda \big| \leq \beta'''_{N_t(q,\sigma)}(\delta) 
\Big\}\,, \quad C^{\text{reward}}_{t,\delta} = \bigcap_{q,\sigma} C^{\text{reward}}_{t,\delta}(q ,\sigma),
}
where $\widehat \nu_{t}(q,\sigma)$ denotes the empirical mean reward built using $N_t(q,\sigma)$ observations collected from the reward distribution $\nu(q,\sigma)$. Here, for $n\!\in\!\NN$, $\beta_n(\delta) = \sqrt{\frac{1}{2n}\big(1+\frac{1}{n}\big)\log\big(\sqrt{n+1}/{\delta}\big)}.$ Then, it suffices to replace $\overline \nu$ with its  upper confidence set, that is, to replace $\overline \nu(q,\sigma)$, in problem (\ref{eq:im-t}), with 
$
\widehat \nu_{t}(q,\sigma)  + \beta'''_{N_t(q,\sigma)}\big(\tfrac{\delta}{Q\Sigma}\big)\, .
$
Further, the parameter $\delta$ in other confidence sets must be rescaled accordingly to account for $C^{\text{reward}}$. Overall, this modification would increase the regret bound by an additive term that is independent of any diameter-like quantity (i.e., $\Dcp$ or $D_{q,o}$). The regret bound will depend on $\sqrt{\log Q}$, which will however be dominated by other $\sqrt{\log T}$ terms. 

\section{Regret Analysis of \UCRLSA}\label{app:regret_proofs_PRM}
In this section, we provide regret analyses of the two variants of \UCRLSA.

\subsection{Proof of Theorem \ref{thm:regret_UCRLSAL1}}

As in most regret analyses for model-based algorithms that work based on the optimism principle, the proof builds on the regret analysis by  \citet{jaksch2010near}, but it includes novel steps due to the structure of MDPRMs. 

Let $\delta\in (0,1)$. We closely follow the notations used by \citet{jaksch2010near}. To simplify notations, we define the short-hand $J_k:=J_{t_k}$ for various random variables that are fixed within a given episode $k$ and omit their dependence on $\delta$ (for example $\cM_{k}:=\cM_{t_k,\delta}$). We let $m(T)$ denote the number of episodes initiated by the algorithm up to time $T$.

Observe that $\EE[r_t|s_t,a_t] = \overline{\nu}(q_t,L(o_t,a_t))$. Hence, by applying Corollary S1 in the online companion, we deduce that
	\als{
		\kR(T)&= \sum_{t=1}^T g^\star- \sum_{t=1}^T r_t \\
		&\leq \sum_{t=1}^T\sum_{o,q,a} \big(g^\star - \overline{\nu}(q,L(o,a))\big)\mathbb I_{\{(q_t,o_t,a_t)=(q,o,a)\}} + \sqrt{ \tfrac{1}{2}(T+1) \log( \sqrt{T+1}/\delta)}\\
		&= \sum_{o,q,a} \big(g^\star - \overline{\nu}(q,L(o,a))\big) N_{m(T)}(q,o,a) + \sqrt{ \tfrac{1}{2}(T+1) \log( \sqrt{T+1}/\delta)}\, ,
	}
	with probability at least $1-\delta$.
	
For $s=(q,o)$, define
$
\mu(s,a):=\overline{\nu}(q,L(o,a)) 
$. 
Hence, the first term in the previous inequality reads
	\als{
		\sum_{s,a} (g^\star - \mu(s,a))N_{m(T)}(s,a) &= \sum_{k=1}^{m(T)}  \sum_{s,a} \underbrace{\sum_{t=t_k}^{t_{k+1}-1}\mathbb I_{\{s_t=s,a_t=a\}} }_{:=\counter_k(s,a)}\big(g^\star - \mu(s,a)\big)=
		\sum_{k=1}^{m(T)} \sum_{s,a} \counter_k(s, a) \big(g^\star - \mu(s, a)\big)\,.
	}
	Introducing $\Delta_k := \sum_{s,a}\counter_k(s,a) \big(g^\star - \mu(s, a)\big)$ for $1\leq k\leq m(T)$, we get
	\als{
		\kR(T) \le \sum_{k=1}^{m(T)} \Delta_k + \sqrt{ \tfrac{1}{2}(T+1) \log( \sqrt{T+1}/\delta)}\, , \qquad \hbox{with probability at least $1-\delta$.}
	}
 	A given episode $k$ is called \emph{good} if $M \in \cM_{k}$, and \emph{bad} otherwise.
	
	\subsubsection{Control of the regret due to bad episodes}
	As a consequence of Lemma \ref{lem:CI_has_true_MDPRM} and Lemma \ref{lem:inclusion_model_sets}, the set $\widetilde \cM^{\times,1}_{k}$ contains the cross-product MDP $\cpM$ associated to the true MDPRM $M$ with probability higher than $1-\delta$ uniformly for all $T$, and for all episodes $k=1,\ldots,m(T)$. As a consequence, with probability at least $1-\delta$, 
	$$
	\sum_{k=1}^{m(T)}\Delta_k\mathbb I_{\big\{\cpM \notin \widetilde \cM^{\times,1}_{k}\big\}} = 0.
	$$
	
	\subsubsection{Control of the regret due to good episodes} To upper bound regret in good episodes, we closely follow \citep{jaksch2010near} and decompose the regret to control the transition and reward functions. Consider a good episode $k$. Let $\widetilde M_{k} = (\cO,
\cA,\widetilde P_k,{\widetilde \cR}_k,{\AtomProp},L)$ denote the chosen optimistic MDPRM at episode $k$.    
    Since $\cpM\in \widetilde \cM^{\times,1}_{k}$, we have $g_k := g^{\pi_k}(\widetilde M_{k}) \geq g^\star - \frac{1}{\sqrt{t_k}}$. Hence, the regret accumulated in episode $k$ satisfies:
	\begin{align}
	\Delta_k  &\leq \sum_{s,a} \counter_k(s,a)
	\big(g_k - \mu(s, a)\big) + \sum_{s,a} \frac{\counter_k(s,a)}{\sqrt{t_k}}
	\, .
	\label{eq:delta_init}
	\end{align}

As a result of \citep[Theorem~8.5.6]{puterman2014markov}, when the convergence
criterion of \EVI\ holds at iterate $i$, we have 
\begin{equation}
\big|u_k^{(i+1)}(s) - u_k^{(i)}(s) - g_k\big| \leq \frac{1}{\sqrt{t_k}}\,, \qquad  \forall s \in \cQ\times \cO.
\label{eq:EVI_convergence}
\end{equation}
By the design of \EVI, note  that for all $s\in \cQ\times \cO$,
	\als{
		u_k^{(i+1)}(s) = \mu(s,\pi_k(s)) + \sum_{s'\in \cQ\times \cO} \widetilde P_k(o'|o,\pi_{k}(s))\widetilde \tau(q'|q,\sigma_k(s)) u_k^{(i)}(s')\, ,
	}
    with $\sigma_k(s):=L(o,\pi_k(s))$, and 
	where we recall that $\widetilde P_k$ and $\widetilde \tau_k$ are the transition probability distribution and RM state distribution of the optimistic MDPRM $\widetilde{M}_k$ in $s=(q,o)$, respectively. 
	Then, (\ref{eq:EVI_convergence}) gives, for all $s\in \cS$,
	\begin{equation*}
	\Big|g_k -    \mu(s,\pi_{k}(s))  - \Big( \sum_{s'} \widetilde P_k(o'|o,\pi_{k}(s)) \widetilde \tau_k(q'|q,\sigma_k(s))u_k^{(i)}(s') - u_k^{(i)}(s)\Big)\Big|
	\leq \frac{1}{\sqrt{t_k}}\, .
	\end{equation*}
	Defining
	${\bf g}_k = g_k \mathbf 1$, ${{\boldsymbol{\mu}}}_k := \big(\mu(s,\pi_k(s))\big)_s$, $\widetilde{\mathbf{P}}^\times_k := \Big(\widetilde P_{k}\big(o'|o,\pi_k(s)\big)\widetilde \tau_k(q'|q,\sigma_k(s))\Big)_{s, s'}$, and $\mathbf\counter_{k} := \big(\counter_k\big(s,\pi_k(s)\big)\big)_s$, we can rewrite the above inequality as:
	\begin{equation*}
	\Big|{\bf g}_k - {\boldsymbol{\mu}}_k - (\widetilde{\mathbf{P}}^\times_k - \mathbf{I}) u_k^{(i)} \Big|
	\leq \frac{1}{\sqrt{t_k}}\mathbf 1\,.
	\end{equation*}
	Also, we can rewrite (\ref{eq:delta_init}) as
	\begin{align}
	\Delta_k
	\leq \sum_{s,a} \counter_k(s,a) \big(g_k-\mu(s,a)\big) + \sum_{s,a} \frac{\counter_k(s,a)}{\sqrt{t_k}}
    &\leq \mathbf\counter_{k} (\widetilde{\mathbf{P}}^\times_k - \mathbf{I} ) u_k^{(i)}  + 2\sum_{s,a} \frac{\counter_k(s,a)}{\sqrt{N_k(o,a)}} \,,
    \label{eq:Delta_k_EVI}
	\end{align}
where in the inequality we used that $t_k\ge \max_{o,a} N_k(o,a)$ so that 
\als{
\sum_{s,a}  \frac{\counter_k(s,a)}{\sqrt{t_k}} \le \sum_{o,a} \frac{1}{\sqrt{N_k(o,a)}} \sum_{q}\counter_k(q,o,a) = \sum_{o,a} 	\frac{\counter_k(o,a)}{\sqrt{N_k(o,a)}}\,. 
}

Let us define, for all $s\in \cQ\times \cO$, 
$$
w_k(s) := u_k^{(i)}(s) - \frac{1}{2}\Big(\min_{s' \in \cS} u_k^{(i)}(s') + \max_{s' \in \cS} u_k^{(i)}(s')\Big)\, .
$$
In view of the fact that $\widetilde{\mathbf{P}}^\times_k$ is row-stochastic (i.e., its rows sum to one), we obtain
	\begin{align}
	\label{eq:Delta_k_2}
	\Delta_k &\leq \mathbf\counter_k (\mathbf{P}^\times_k-\mathbf{I})w_k + \mathbf\counter_k (\widetilde{\mathbf{P}}^\times_k - \mathbf{P}^\times_k) w_k +  2\sum_{o,a} 	\frac{\counter_k(o,a)}{\sqrt{N_k(o,a)}} \, .	
	\end{align}

The following lemmas control the first and second terms in the right-hand side of (\ref{eq:Delta_k_2}):

\begin{lemma}
    \label{lem:Ptilde_P_diff_UB}
    For any good episode $k$, we have
    \begin{align*}
            \mathbf\counter_k (\widetilde{\mathbf{P}}^\times_k - \mathbf{P}^\times_k) w_k &\le 2\Dcp\sqrt{\log\big(2OA\sqrt{T}/\delta\big) + O}\sum_{a\in \cA}\sum_{o\in \cO} \frac{\counter_k(o,a)}{\sqrt{N_k(o,a)}} \\
            &+ 2\Dcp\sqrt{\log\big(2\sum_q E_q\sqrt{T}/\delta\big) + Q}\sum_{q\in \cQ}\sum_{\sigma\in\cE_q} \frac{\counter_k(q,\sigma)}{\sqrt{N_k(q,\sigma)}} \, .
    \end{align*}
\end{lemma}

\begin{lemma}
    \label{lem:P_I_diff_UB}
    Uniformly over all $T\ge 1$, with probability exceeding $1-\delta$, we have
    \begin{align*}
            \sum_{k=1}^{m(T)} \mathbf\counter_k (\mathbf{P}^\times_k - \mathbf I) w_k \mathbb I_{\big\{\cpM \in \widetilde \cM^{\times,1}_{k}\big\}} \leq \Dcp \sqrt{\tfrac{1}{2}(T+1) \log(\sqrt{T+1}/\delta)} + \Dcp OA\log_2\big(\tfrac{8T}{OA}\big) + \Dcp QE\log_2\big(\tfrac{8T}{QE}\big)\, .
    \end{align*}
\end{lemma}

Using the bounds in Lemma \ref{lem:Ptilde_P_diff_UB}) and Lemma \ref{lem:P_I_diff_UB} and summing over all good episodes, for the regret built during the good episodes we obtain:
	\begin{align}
	\sum_{k=1}^{m(T)}  \Delta_k \mathbb I_{\big\{\cpM \in \widetilde \cM^{\times,1}_{k}\big\}} &\le  2\big(\Dcp+2\big)\sqrt{\log\big(2OA\sqrt{T}/\delta\big) + O}\sum_{k=1}^{m(T)} \sum_{a\in \cA}\sum_{o\in \cO} \frac{\counter_k(o,a)}{\sqrt{N_k(o,a)}} \sk
    &+ 2\Dcp\sqrt{\log\big(2QE\sqrt{T}/\delta\big) + Q}\sum_{k=1}^{m(T)} \sum_{q\in \cQ}\sum_{\sigma\in\cE_q} \frac{\counter_k(q,\sigma)}{\sqrt{N_k(q,\sigma)}} \sk
    &+ \Dcp \sqrt{\tfrac{1}{2}(T+1) \log(\sqrt{T+1}/\delta)} + \Dcp (OA + QE)\log(T)\,,
	\label{eq:out_together}
	\end{align}
with probability exceeding $1-\delta$. 

\subsubsection{The Final Bound} To derive the final bound, we simply and bound the right-hand side of (\ref{eq:out_together}) as follows. 
Applying Lemma \ref{lem:seq_sqrt} and using the Cauchy-Schwarz inequality:
\begin{align*}
    \sum_{k=1}^{m(T)} \sum_{o,a} \frac{\counter_k(o,a)}{\sqrt{N_k(o,a)}} &\le 3\sum_{o, a}\sqrt{N_T(o, a)} \le 3\sqrt{\sum_{o, a} N_T(o, a) } = 3\sqrt{OAT} \, ,
\end{align*}
where, with a slight abuse of notation, we used $N_T(o,a)$ to denote the number of visits to $(o,a)$ after $T$ rounds. Similarly, we have
\begin{align*}
    \sum_{k=1}^{m(T)} \sum_{q,\sigma\in \cE_q} \frac{\counter_k(q,\sigma)}{\sqrt{N_k(q,\sigma)}} &\le 3\sum_{q, \sigma\in \cE_q}\sqrt{N_T(q, \sigma)} \le 3\sqrt{\sum_{q} E_q \sum_{q, \sigma\in\cE_q} N_T(q, \sigma) } = 3\sqrt{\sum_{q} E_q T} \, .
\end{align*}

Combining these with (\ref{eq:out_together}), we have that with probability at least $1-4\delta$,
	\als{
		\kR(T) &\leq  6\Dcp\sqrt{\log\big(2OA\sqrt{T}/\delta\big) + O}\sqrt{OAT}+ 6\Dcp\sqrt{\log\big(2QE\sqrt{T}/\delta\big) + Q}\sqrt{\sum_{q} E_q T} \sk
    &+ (\Dcp + 1) \sqrt{\tfrac{1}{2}(T+1) \log(\sqrt{T+1}/\delta)} 	+ \Dcp (OA + QE)\log(T)\, ,
	}
thus proving the theorem. 
\ep

\subsection{Proof of Theorem \ref{thm:regret_UCRLSAB}}

\bp
Let $\delta\in (0,1)$. Following the same steps as in the proof of Theorem \ref{thm:regret_UCRLSAL1}, we have
	\als{
		\kR(T) \le \sum_{k=1}^{m(T)} \Delta_k + \sqrt{ \tfrac{1}{2}(T+1) \log( \sqrt{T+1}/\delta)}\, ,
	}
	with probability at least $1-\delta$, where $\Delta_k$ is defined similarly to the proof of Theorem \ref{thm:regret_UCRLSAL1}. 
    
As a consequence of Lemma \ref{lem:CI_has_true_MDPRM} and Lemma \ref{lem:inclusion_model_sets}, the set $\widetilde \cM^{\times,2}_{k}$ contains the cross-product MDP $\cpM$ associated to the true MDPRM $M$ with probability higher than $1-\delta$ uniformly for all $T$, and for all episodes $k=1,\ldots,m(T)$. As a consequence, with probability at least $1-\delta$, 
	$$
	\sum_{k=1}^{m(T)}\Delta_k\mathbb I_{\big\{\cpM \notin \widetilde \cM^{\times,2}_{k}\big\}} = 0.
	$$

Let us now focus on good episodes, i.e., episodes $k$ where $M\in \cM_k$. Similarly to the proof of Theorem \ref{thm:regret_UCRLSAL1}, we have: 
	\begin{align*}
	\Delta_k  &\leq \sum_{s,a} \counter_k(s,a)
	\big(g_k - \mu(s, a)\big) + \sum_{s,a} \frac{\counter_k(s,a)}{\sqrt{t_k}} \leq \mathbf\counter_k (\mathbf{P}^\times_k-\mathbf{I})w_k + \mathbf\counter_k (\widetilde{\mathbf{P}}^\times_k - \mathbf{P}^\times_k) w_k +  2\sum_{o,a} 	\frac{\counter_k(o,a)}{\sqrt{N_k(o,a)}}\, ,
	\end{align*}
where $w_k$ is defined in the proof of Theorem \ref{thm:regret_UCRLSAL1}. The first term above is controlled via Lemma \ref{lem:P_I_diff_UB}. To bound the second term, we use the following lemma: 
\begin{lemma}
    \label{lem:Ptilde_P_diff_UB_Bernstein}
    For any good episode $k$, we have
    \begin{align*}
            \mathbf\counter_k &(\widetilde{\mathbf{P}}^\times_k - \mathbf{P}^\times_k) w_k \le \frac{\Dcp}{2}\Bigg(\sum_{o,a} n_k(o,a)\sqrt{\frac{2K_{o,a}\ell}{N_k(o,a)}} + \sum_{q,\sigma}  \counter_k(q,\sigma)\sqrt{\frac{2K_{q,\sigma}\ell}{N_k(q,\sigma)}} \\
    &+  4O\ell\sum_{o,a}  \frac{\counter_k(o,a)}{N_k(o,a)}  + 4QO\ell\sum_{q,\sigma}  \frac{\counter_k(q,\sigma)}{N_k(q,\sigma)}   +  O\ell\sqrt{8Q}\sum_{q,o,a}  \frac{\counter_k(q,o,a)}{N_k(q,o,a)} + 6OQ\ell^{3/2}\sum_{o,a}  \frac{\counter_k(o,a)}{N_k(o,a)^{3/2}}\Bigg)
    \,,
    \end{align*}
   with $\ell = \cO(\log(\log(T)/\delta) + \log(OQA))$.
\end{lemma}

Applying Lemmas \ref{lem:Ptilde_P_diff_UB_Bernstein} and \ref{lem:P_I_diff_UB}, and using similar algebraic manipulations as in the proof of Theorem \ref{thm:regret_UCRLSAL1} yield the desired bound:
\als{
		\kR(T) &\leq  3\Dcp \sqrt{\sum_{o,a}K_{o,a}T\ell} + 3\Dcp \sqrt{\sum_{q,\sigma}K_{q,\sigma}T\ell}  + (\Dcp + 1) \sqrt{\tfrac{1}{2}(T+1) \log(\sqrt{T+1}/\delta)} 	+ c\log(T)\ell\, ,
	}
with $c=\Dcp (OA + QE + 4O^2A + 4OQ^2E + QO^2A + Q^2O^2A)$.    
\ep

\subsection{Technical Lemmas}
\label{sec:supporting_lemmas}

\subsubsection{Proof of Lemma \ref{lem:Ptilde_P_diff_UB}}

\bp
 We have	
	\begin{align*}
	\mathbf\counter_k (\widetilde{\mathbf{P}}^\times_k - \mathbf{P}^\times_k) w_k &= \sum_{s\in \cS} \sum_{s'\in \cS} \counter_k(s,\pi_k(s))  \Big(\widetilde P^\times_k(s'|s,\pi_k(s)) - P^\times(s'|s,\pi_k(s))\Big) w_k(s') \\
	&\le \|w_k\|_\infty  \sum_{s\in \cS} \counter_k(s,\pi_k(s)) \Big\|\widetilde P^\times_k(\cdot|s,\pi_k(s)) - P^\times(\cdot|s,\pi_k(s))\Big\|_1 \\
    &\le \Dcp  \sum_{s\in \cS} \counter_k(s,\pi_k(s)) \Big\|\widetilde P^\times_k(\cdot|s,\pi_k(s)) - P^\times(\cdot|s,\pi_k(s))\Big\|_1 \\
    &\le 2\Dcp  \sum_{s\in \cS} \counter_k(s,\pi_k(s)) \Big(\beta_{N_k(o,\pi_k(s))} + \beta''_{N_k(q,\sigma_k(s))}\Big)\,,
 \end{align*}
where we used the fact that $\|w_k\|_\infty \le \Dcp/2$, as established by \cite{jaksch2010near} for communicating MDPs. 

Observe that 
\begin{align*}
    \sum_{s\in \cS} \counter_k(s,\pi_k(s)) \beta_{N_k(o,\pi_k(s))}&\le \sum_{a\in \cA}\sum_{q\in \cQ}\sum_{o\in \cO} \counter_k(q,o,a) \beta_{N_k(o,a)} \\
    &\le \sum_{a\in \cA}\sum_{o\in \cO} \beta_{N_k(o,a)}\sum_{q\in \cQ}\counter_k(q,o,a)  \\
    &= \sum_{a\in \cA}\sum_{o\in \cO} \counter_k(o,a) \beta_{N_k(o,a)} \\
    &\le 2\sqrt{\log\big(2OA\sqrt{T}/\delta\big) + O}\sum_{a\in \cA}\sum_{o\in \cO} \frac{\counter_k(o,a)}{\sqrt{N_k(o,a)}}\,,
\end{align*}
where the last inequality follows from the definition of $\beta_{N_k(o,a)}$ together with $1\le N_k(o,a)\le T$.

Furthermore, 
\begin{align*}
    \sum_{s\in \cS} \counter_k(s,\pi_k(s)) \beta''_{N_k(q,\sigma_k(s))} &\le \sum_{q\in \cQ}\sum_{o\in \cO}\sum_{a\in \cA} \counter_k(q,o,a) \beta''_{N_k(q,L(o,a))} \\
    &\le \sum_{q\in \cQ}\sum_{\sigma\in \cE_q} \counter_k(q,\sigma) \beta''_{N_k(q,\sigma)} \\
    &\le 2\sqrt{\log\big(2QE\sqrt{T}/\delta\big) + Q}\sum_{q\in \cQ}\sum_{\sigma\in \cE_q} \frac{\counter_k(q,\sigma)}{\sqrt{N_k(q,\sigma)}}\,,
\end{align*}
where the second inequality uses the observation that a visit to $(o,a)$ implies collecting $\sigma$. Further, the last inequality follows from the definition of $\beta''_{N_k(q,\sigma)}$ together with $1\le N_k(q,\sigma)\le T$. Putting together proves the lemma.\ep

\subsubsection{Proof of Lemma \ref{lem:P_I_diff_UB}}

\bp
    Similarly to the proof of Theorem 2 in \citep{jaksch2010near}, we define a martingale difference sequence $(Z_t)_{t\geq 1}$ with 
$$
Z_t := (P^\times_k(\cdot|s_t,a_t) - \mathbf e_{s_{t+1}})w_{k(t)}\indic{M \in \cM_{k(t)}},
$$
for all $t$, where $k(t)$ denotes the episode containing time step $t$. For any good episode $k$, we have:
	\als{
		\mathbf\counter_k (\mathbf{P}^\times_k-\mathbf{I})w_k  &= \sum_{t=t_k}^{t_{k+1} -1} (P^\times_k(\cdot|s_t,a_t) - \mathbf{e}_{s_t}) w_k = \sum_{t=t_k}^{t_{k+1} -1} \Big(P^\times_k(\cdot|s_t,a_t) - \textbf{e}_{s_{t+1}} + \textbf{e}_{s_{t+1}} - \textbf{e}_{s_t} \Big) w_k 	\\
		&= \sum_{t=t_k}^{t_{k+1} -1} Z_t + w_k(s_{t_{k+1}}) - w_k(s_{t_k}) \leq \sum_{t=t_k}^{t_{k+1} -1} Z_t + \Dcp \, ,
	}
	where $\mathbf e_i$ denotes a vector with the $i$-th element being $1$ and the others being zero. Hence, $\sum_{k=1}^{m(T)} \counter_k (\mathbf{P}^\times_k-\mathbf{I})w_k \leq \sum_{t=1}^T Z_t +m(T)\Dcp$. As established in \citep{jaksch2010near}, $|Z_t|\le \|P^\times_k(\cdot|s_t,a_t) - \textbf{e}_{s_{t+1}} \|_1\|w_{k(t)}\|_\infty\le \Dcp$ and $\mathbb{E}[Z_t|s_1, a_1, \dots, s_t, a_t] = 0$, so that $(Z_t)_{t\ge 1}$ is martingale difference sequence. Therefore, by  Corollary S1, we get:
	\begin{align*}
	\mathbb P\bigg(\forall  T: \sum_{t=1}^T Z_t  \leq \Dcp \sqrt{\tfrac{1}{2} (T+1) \log(\sqrt{T+1}/\delta)}\bigg)\geq 1-\delta\,.
	\end{align*}
Putting together with the bound on $m(T)$ in Lemma \ref{lem:no_episodes_UB} concludes the proof.%
\ep

\subsubsection{Proof of Lemma \ref{lem:Ptilde_P_diff_UB_Bernstein}}

\bp 
We have 
	\begin{align*}
	\mathbf\counter_k (\widetilde{\mathbf{P}}^\times_k - \mathbf{P}^\times_k) w_k &= \sum_{s\in \cS} \sum_{s'\in \cS} \counter_k(s,\pi_k(s))  \Big(\widetilde P^\times_k(s'|s,\pi_k(s)) - P^\times(s'|s,\pi_k(s))\Big) w_k(s') \\
	&\le \frac{\Dcp}{2}\sum_{s\in \cS} \sum_{s'\in \cS} \counter_k(s,\pi_k(s))  \Big|\widetilde P^\times_k(s'|s,\pi_k(s)) - P^\times(s'|s,\pi_k(s))\Big| \,.
 \end{align*}

Fix $s=(q,o)$. In view of the definition of $\cP^2$, we have
\begin{align}
    \sum_{s'\in \cS} \Big|\widetilde P^\times_k(s'|s,\pi_k(s)) - P^\times(s'|s,\pi_k(s))\Big|  
    &\le \sum_{q',o'} \widehat \tau_t(q'|q,\sigma)\sqrt{\frac{2u_{o,a,o'}(1-u_{o,a,o'})}{N_k(o,a)}\ell} \sk 
    &+ \sum_{q',o'} u_{o,a,o'}\sqrt{\frac{2\lambda_{q,\sigma,q'}(1-\lambda_{q,\sigma,q'})}{N_k(q,\sigma)}\ell} \sk
    &+ \sum_{q',o'} \Big(\frac{\widehat \tau_t(q'|q,\sigma)\ell}{3N_k(o,a)} + \frac{u_{o,a,o'}\ell}{3N_k(q,\sigma)}\Big) \,, \label{eq:tmp1}
\end{align}
with $\sigma = L(o,a)$. 
Note that $\beta'_{N_k} = \cO(\log(\log(T)/\delta))$. We use the short-hand $\ell$ to upper bound both $\beta'_{N_k(o,a)}\big(\frac{\delta}{4O^2A}\big)$ and $\beta'_{N_k(q,\sigma)}\big(\frac{\delta}{4Q^2E}\big)$.  

To upper bound the first term in the right-hand side of (\ref{eq:tmp1}), first observe that  
\begin{align*}
    \sum_{o'} \sqrt{u_{o,a,o'}(1-u_{o,a,o'})} &\le \sum_{o'} \sqrt{\widehat P_k(o'|o,a)(1-\widehat P_k(o'|o,a))} + \sum_{o'} 3 \sqrt{\frac{\ell}{N_k(o,a)}} \\
    &\le \sum_{o'} \sqrt{\widehat P_k(o'|o,a)(1-\widehat P_k(o'|o,a))} + 3O\sqrt{\frac{\ell}{N_k(o,a)}} \\
    &\le \sqrt{K_{o,a}-1} + 3O\sqrt{\frac{\ell}{N_k(o,a)}}\,,
\end{align*}
where the first inequality is due to Lemma \ref{lem:SqrtVar_pqBern}, whereas the last uses the fact that for a distribution $p\in \Delta_\cO$ with $K$ non-zero elements, we have
\als{
\sum_{o\in \cO} \sqrt{p(o)(1-p(o))}= \sum_{o:p(o)>0} \sqrt{p(o)(1-p(o))} \sqrt{\sum_{o:p(o)>0} p(o) \sum_{o:p(o)>0} (1-p(o))} = \sqrt{K-1}\,.
}
Thus, 
\begin{align}
    \sum_{q',o'} \widehat \tau_t(q'|q,\sigma)\sqrt{\frac{2u_{o,a,o'}(1-u_{o,a,o'})}{N_k(o,a)}\ell} &= \sum_{o'} \sqrt{\frac{2u_{o,a,o'}(1-u_{o,a,o'})}{N_k(o,a)}\ell} \sum_{q'}\widehat \tau_t(q'|q,\sigma) \sk
    &\le \sqrt{\frac{2K_{o,a}\ell}{N_k(o,a)}} +  \frac{3O\ell}{N_k(o,a)} \,.\label{eq:UB_1st}
\end{align}
We use a similar technique to bound the second term in (\ref{eq:tmp1}): 
\begin{align}
    \sum_{q',o'} u_{o,a,o'}&\sqrt{\frac{2\lambda_{q,\sigma,q'}(1-\lambda_{q,\sigma,q'})}{N_k(q,\sigma)}\ell} \le  \bigg(\sqrt{\frac{K_{q,\sigma}\ell}{N_k(q,\sigma)}} +  \frac{3Q\ell}{N_k(q,\sigma)}\bigg) \sum_{o'}u_{o,a,o'}  \sk
    &\stackrel{\text{(i)}}{\le} \bigg(\sqrt{\frac{2K_{q,\sigma}\ell}{N_k(q,\sigma)}} +  \frac{3Q\ell}{N_k(q,\sigma)}\bigg)\bigg(1 + 2O\sqrt{\frac{\ell}{N_k(o,a)}}\bigg)  \sk
    &\le \sqrt{\frac{K_{q,\sigma}\ell}{N_k(q,\sigma)}} +  \frac{3Q\ell}{N_k(q,\sigma)}
    +  O\ell\sqrt{\frac{8Q}{N_k(o,a)N_k(q,\sigma)}} + \sqrt{\frac{\ell}{N_k(o,a)}}\frac{6OQ\ell}{N_k(q,\sigma)} \sk
    &\stackrel{\text{(ii)}}{\le} \sqrt{\frac{2K_{q,\sigma}\ell}{N_k(q,\sigma)}} +  \frac{3Q\ell}{N_k(q,\sigma)}
    +  \frac{O\ell \sqrt{8Q}}{N_k(q,o,a)} + \frac{6OQ\ell^{3/2}}{N_k(o,a)^{3/2}} 
    \,,\label{eq:UB_2nd}
\end{align}
where (i) follows from the following inequality 
\begin{align*}
    \sum_{o'}u_{o,a,o'} &\le \sum_{o'}\widehat P_k(o'|o,a) + \sum_{o'}\sqrt{\frac{2u_{o,a,o'}(1-u_{o,a,o'})}{N_k(o,a)} \ell} + \sum_{o'}\frac{\ell}{3N_k(o,a)} \\
    &\le 1 + O\sqrt{\frac{\ell}{2N_k(o,a)}} +  \frac{O\ell}{3N_k(o,a)} \le 1 + 2O\sqrt{\frac{\ell}{N_k(o,a)}}
\, ,
\end{align*}
and where (ii) uses that $N_k(q,\sigma)\ge N_k(q,o,o)$ and $N_k(o,a)\ge N_k(q,o,o)$.

Finally, the last term in (\ref{eq:tmp1}) is bounded as
\begin{align*}
    \sum_{q',o'} \Big(\frac{\widehat \tau_t(q'|q,\sigma)\ell}{3N_k(o,a)} + \frac{u_{o,a,o'}\ell}{3N_k(q,\sigma)}\Big) &\le \frac{\ell}{3N_k(o,a)} \sum_{o',q'} \widehat \tau_t(q'|q,\sigma) 
    + \frac{\ell}{3N_k(q,\sigma)} \sum_{q',o'} u_{o,a,o'} \\
    &\le \frac{O\ell}{3N_k(o,a)} + \frac{QO\ell}{3N_k(q,\sigma)}\, .
\end{align*}

Putting this together with (\ref{eq:UB_1st}) and (\ref{eq:UB_2nd}) gives the desired bound: 
\begin{align*}
    \mathbf\counter_k &(\widetilde{\mathbf{P}}^\times_k - \mathbf{P}^\times_k) w_k \le \frac{\Dcp}{2}\sum_{q,o,a} \counter_k(q,o,a) \bigg(\sqrt{\frac{2K_{o,a}}{N_k(o,a)}\ell} +  \frac{3O\ell}{N_k(o,a)} + \frac{O\ell}{3N_k(o,a)} + \frac{QO\ell}{3N_k(q,\sigma)} \\
    &\qquad + \sqrt{\frac{2K_{q,\sigma}}{N_k(q,\sigma)}\ell} +  \frac{3Q\ell}{N_k(q,\sigma)}
    +  \frac{O\ell\sqrt{8Q}}{N_k(q,o,a)} + \frac{6OQ\ell^{3/2}}{N_k(o,a)^{3/2}}\bigg)\\
    &\le \frac{\Dcp}{2}\Bigg(\sum_{o,a} n_k(o,a)\sqrt{\frac{2K_{o,a}\ell}{N_k(o,a)}} + \sum_{q,\sigma}  \counter_k(q,\sigma)\sqrt{\frac{2K_{q,\sigma}\ell}{N_k(q,\sigma)}} \\
    &+  4O\ell\sum_{o,a}  \frac{\counter_k(o,a)}{N_k(o,a)}  + 4QO\ell\sum_{q,\sigma}  \frac{\counter_k(q,\sigma)}{N_k(q,\sigma)}   +  O\ell\sqrt{8Q}\sum_{q,o,a}  \frac{\counter_k(q,o,a)}{N_k(q,o,a)} + 6OQ\ell^{3/2}\sum_{o,a}  \frac{\counter_k(o,a)}{N_k(o,a)^{3/2}}\Bigg)
    \,.
\end{align*}
\ep

\subsubsection{Auxiliary Lemmas}

\begin{lemma}[{\citep[Lemma 11]{bourel2020tightening}}]
	\label{lem:SqrtVar_pqBern}
	Consider $x$ and $y$ satisfying $|x - y| \leq \sqrt{2y(1-y)\zeta} + \zeta/3$. Then,
    $
	\sqrt{y(1 - y)} \leq \sqrt{x(1- x)} + 2.4\sqrt{\zeta}
	$.
\end{lemma}

\begin{lemma}[{\citep[Lemma~19]{jaksch2010near},\citep[Lemma~24]{talebi2018variance}}]\label{lem:seq_sqrt}
	Consider $(z_i)_{1\le i\le n}$ with $0 \leq z_k \leq Z_{k-1} := \max\{1, \sum_{i=1}^{k-1}z_i\}$. Then: (i) $\sum_{k=1}^n 	\frac{z_k}{\sqrt{Z_{k-1}}} 	\leq 	\big(\sqrt{2} + 1\big) 	\sqrt{Z_n}$ and (ii)$\sum_{k=1}^n \frac{z_k}{Z_{k-1}} \leq 2\log(Z_n) + 1$.
\end{lemma}

The following lemma is a straightforward extension of \citep[Proposition~18]{jaksch2010near}:

\begin{lemma}
\label{lem:no_episodes_UB}
The number $m(T)$ of episodes 
up to time $T\ge OA$ satisfies
$
m(T) \le OA\log_2\big(\tfrac{8T}{OA}\big) + QE \log_2\big(\tfrac{8T}{QE}\big) \, .
$
\end{lemma}

\section{Regret Analysis of Deterministic RMs}\label{app:regret_deterministic}
\subsection{Proof of Theorem \ref{thm:regret_UCRLSAL1_deterministic}}

\bp
We can using similar lines as in the proof of Theorem \ref{thm:regret_UCRLSAL1}, 
with slight modifications to account for the fact that $\tau$ is a deterministic function, and is hence assumed known. In doing so, we obtain the following bound on the regret built during a good episode $k$: 
\begin{align}\label{eq:Delta_k_RM_L1}
	\Delta_k &\leq \mathbf\counter_k (\mathbf{P}^\times_k-\mathbf{I})w_k + \mathbf\counter_k (\widetilde{\mathbf{P}}^\times_k - \mathbf{P}^\times_k) w_k +  2\sum_{o,a} 	\frac{\counter_k(o,a)}{\sqrt{N_k(o,a)}} \, ,	
\end{align}
with $w_k$ defined as follows:  
$$
w_k(s) := u_k^{(i)}(s) - \frac{1}{2}\Big(\min_{s' \in \cB_{s} \times \cO} u_k^{(i)}(s') + \max_{s' \in \cB_s \times \cO} u_k^{(i)}(s')\Big)\, ,\quad \forall s\in \cS.
$$
In the case of deterministic RM, we upper bound $\mathbf\counter_k (\widetilde{\mathbf{P}}^\times_k - \mathbf{P}^\times_k) w_k$ as follows: 
	\begin{align}
	\mathbf\counter_k (\widetilde{\mathbf{P}}^\times_k &- \mathbf{P}^\times_k) w_k = \sum_{s\in \cS} \sum_{s'\in \cS} \counter_k(s,\pi_k(s))  \Big(\widetilde P^\times_{k}(s'|s,\pi_k(s)) - P^\times(s'|s,\pi_k(s))\Big) w_k(s') \sk
	&= \sum_{s\in \cS} \counter_k(s,\pi_k(s)) \sum_{o'\in \cO} \sum_{q'\in \cQ}  \Big(\widetilde P_{k}(o'|o,\pi_k(s)) - P(o'|o,\pi_k(s))\Big)\indic{q' = \tau(q, \eventFunction(o, \pi_k(s)))} w_k(q',o') \sk
	&\stackrel{\text{(i)}}{\le} \sum_{s\in \cS} \counter_k(s,\pi_k(s)) \sum_{o'\in \cO}   \Big|\widetilde P_k(o'|o,\pi_k(s)) - P(o'|o,\pi_k(s)) \Big|\sk
    &\qquad\qquad\qquad \times \max_{s' \in \cB_{q, o} \times \cO}\big|w_k(q',o')\big| \underbrace{\sum_{q'\in \cQ}\indic{q' = \tau(q, \eventFunction(o, \pi_k(s)))}}_{=1} \sk
	&\le \sum_{s\in \cS} \counter_k(s,\pi_k(s)) \big\|\big(\widetilde P_{k}-P\big)(\cdot|o,\pi_k(s))\big\|_1\cdot \max_{s' \in \cB_{q, o} \times \cO}\big|w_k(q',o')\big|  \sk
	&\stackrel{\text{(ii)}}{\le} \sum_{o\in \cO}\sum_{q\in \cQ} \counter_k(q,o,\pi_k(q,o)) \beta_{N_k(o,\pi_k(q,o))}\big(\tfrac{\delta}{OA}\big) \cdot D_{q,o} \sk
	&\le \sum_{a\in \cA}\sum_{o\in \cO}\sum_{q\in \cQ} \counter_k(q,o,a) \cdot \beta_{N_k(o,a)}\big(\tfrac{\delta}{OA}\big) \cdot D_{q,o}	\sk
	&\le \sum_{a\in \cA}\sum_{o\in \cO}\beta_{N_k(o,a)}\big(\tfrac{\delta}{OA}\big)\cdot \max_{q\in \cQ} D_{q,o}\sum_{q\in \cQ} \counter_k(q,o,a) \sk
	\label{eq:Ptilde_P_diff_RM_L1}
	&\le \sum_{a\in \cA}\sum_{o\in \cO}\beta_{N_k(o,a)}\big(\tfrac{\delta}{OA}\big)\cdot \max_{q\in \cQ} D_{q,o} \cdot \counter_k(o,a) \, ,
	\end{align}
where in (i) we used the definition of $\cB_s$, and where (ii) follows from Lemma \ref{lem:RM_restricted_value_bound}, which is stated and proven at the end of this subsection. Now, combining (\ref{eq:Ptilde_P_diff_RM_L1}) with (\ref{eq:Delta_k_RM_L1}), summing over all good episodes, and applying Lemma 5, 
we obtain:
\begin{align*}
	\sum_{k=1}^{m(T)} \Delta_k \mathbb I_{\big\{\cpM \in \widetilde \cM^{\times,1}_{k}\big\}} &\leq \sum_{k=1}^{m(T)} \mathbf\counter_k (\mathbf{P}^\times_k-\mathbf{I})w_k \mathbb I_{\big\{\cpM \in \widetilde \cM^{\times,1}_{k}\big\}} + \sum_{k=1}^{m(T)}\mathbf\counter_k (\widetilde{\mathbf{P}}^\times_k - \mathbf{P}^\times_k) w_k \mathbb I_{\big\{\cpM \in \widetilde \cM^{\times,1}_{k}\big\}} \\
    &+  2\sum_{k=1}^{m(T)}\sum_{o,a} 	\frac{\counter_k(o,a)}{\sqrt{N_k(o,a)}} \\	
    &\le 2\sqrt{O+\log\big(OA\sqrt{T+1}/\delta\big)} \sum_{k=1}^{m(T)} \sum_{o,a} \Big(\max_{q\in \cQ}D_{q,o}+2\Big)\frac{\counter_k(o,a)}{\sqrt{N_k(o,a)}} \sk
	&+ \Dcp \sqrt{\tfrac{1}{2} (T+1) \log(\sqrt{T+1}/\delta)} + \Dcp OA\log(T),
\end{align*}
holding with probability at least $1-\delta$. Applying Lemma 8 
and using Cauchy-Schwarz yield:
\begin{align*}
    \sum_{k=1}^{m(T)} \sum_{o,a} \max_{q\in \cQ}D_{q,o} \frac{\counter_k(o,a)}{\sqrt{N_k(o,a)}} &\le 3\sum_{o, a}\max_{q\in \cQ}D_{q,o}\sqrt{N_T(o, a)} \le 3\sqrt{\sum_{o, a} \max_{q\in \cQ}D_{q,o}^2\cdot  \sum_{o, a} N_T(o, a) } = 3\sqrt{\mathbf c_M AT}\, ,
\end{align*}
where, with a slight abuse of notation, we used $N_T(o,a)$ to denote the number of visits to $(o,a)$ after $T$ rounds. The rest of the proof follows algebraic manipulations similar to those in the proof of Theorem 1. 
\ep

\begin{lemma}
\label{lem:RM_restricted_value_bound}
For all $(s, a)\in \cS\times\cA$, we have: 
\als{
\max_{s' \in \cB_s \times \cO}|w_k(s')| \leq \frac{D_s}{2}\,, \qquad \|w_k\|_\infty\le \frac{\Dcp}{2}\,.
}
\end{lemma}

\emph{Proof (of Lemma \ref{lem:RM_restricted_value_bound})}
The proof is quite similar to the one of Lemma 8 in  \citep{bourel2020tightening}. 
We first show that for all $s_1,s_2\in \cB_s \times \cO$, we have $u^{(i)}_k(s_1) - u^{(i)}_k(s_2) \leq D_s$, which further implies 
$$
 \max_{x\in \cB_s \times \cO}|w_k(x)| \leq \tfrac{D_s}{2}.
$$	
	To prove this, recall that similarly to \citep{jaksch2010near}, we can combine all cross-product MDPs in $\widetilde \cM^{\times,1}_k$ to form a single MDP $\overline \cM^{\times}_k$ with continuous action space $\cA'$. In this extended MDP, in any $s=(q,o)\in \cS$, and for each $a\in \cA$, there is an action in $\cA'$ with mean $\mu(s,a)$ and transition probability ${\widetilde P}^{\times}_k(\cdot|s,a)$ (of the associated $\cpM$) belonging to the maintained confidence sets. Similarly to \citep{jaksch2010near}, we note  that $u^{(i)}_k(s)$ amounts to the total expected $i$-step reward of an optimal non-stationary $i$-step policy starting in state $s$ on the MDP $\overline \cM^{\times}_k$ with the extended action set. The RM-restricted diameter of state $s$ of $\overline \cM^{\times}_k$ is at most $D_s$, since by assumption $k$ is a good episode and hence $\widetilde \cM^{\times,1}_k$ contains the $\cpM$ associated to the true MDPRM $M$, and therefore, the actions of the true MDPRM are contained in the continuous action set of $\overline \cM^{\times}_k$. Now, if there were states $s_1,s_2\in \cB_s \times \cO$ with $u^{(i)}_k(s_1) - u^{(i)}_k(s_2) > D_s$, then an improved value
for $u^{(i)}_k(s_1)$ could be achieved by the following non-stationary policy: First follow a policy that
moves from $s_1$ to $s_2$ most quickly, which takes at most $D_s$ steps on average. Then follow the optimal
$i$-step policy for $s_2$. We thus have $u^{(i)}_k(s_1) \geq u^{(i)}_k(s_2)-D_s$, since at most $D_s$ of the $i$ rewards of the policy for $s_2$ are missed. This is a contradiction, and so the claim follows. The second bound directly follows from the same arguments as in \citep{jaksch2010near}.\ep

\subsection{Proof of Theorem \ref{thm:regret_UCRLSAB_deterministic}}

\bp 
We can using similar lines as in the proof of Theorem \ref{thm:regret_UCRLSAB}, 
with slight modifications to account for the fact that $\tau$ is a deterministic function, and is hence assumed known. In doing so, we obtain the following bound on the regret built during a good episode $k$: 
\begin{align*}
	\Delta_k &\leq \mathbf\counter_k (\mathbf{P}^\times_k-\mathbf{I})w_k + \mathbf\counter_k (\widetilde{\mathbf{P}}^\times_k - \mathbf{P}^\times_k) w_k +  2\sum_{o,a} 	\frac{\counter_k(o,a)}{\sqrt{N_k(o,a)}} \, ,	
\end{align*}
with $w_k$ defined as follows:  
$$
w_k(s) := u_k^{(i)}(s) - \frac{1}{2}\Big(\min_{s' \in \cB_{s} \times \cO} u_k^{(i)}(s') + \max_{s' \in \cB_s \times \cO} u_k^{(i)}(s')\Big)\, ,\quad \forall s\in \cS.
$$
Here, we upper bound $\mathbf\counter_k (\widetilde{\mathbf{P}}^\times_k - \mathbf{P}^\times_k) w_k$ as follows: 
	\begin{align}
	\mathbf\counter_k (\widetilde{\mathbf{P}}^\times_k &- \mathbf{P}^\times_k) w_k = \sum_{s\in \cS} \sum_{s'\in \cS} \counter_k(s,\pi_k(s))  \Big(\widetilde P^\times_{k}(s'|s,\pi_k(s)) - P^\times(s'|s,\pi_k(s))\Big) w_k(s') \sk
	&= \sum_{s\in \cS} \counter_k(s,\pi_k(s)) \sum_{o'\in \cO} \sum_{q'\in \cQ}  \Big(\widetilde P_{k}(o'|o,\pi_k(s)) - P(o'|o,\pi_k(s))\Big)\indic{q' = \tau(q, \eventFunction(o, \pi_k(s)))} w_k(q',o') \sk
	&\stackrel{\text{(i)}}{\le} \sum_{s\in \cS} \counter_k(s,\pi_k(s)) \sum_{o'\in \cO}   \Big|\widetilde P_k(o'|o,\pi_k(s)) - P(o'|o,\pi_k(s)) \Big|\sk
    &\qquad\qquad\qquad \times \max_{s' \in \cB_{q, o} \times \cO}\big|w_k(q',o')\big| \underbrace{\sum_{q'\in \cQ}\indic{q' = \tau(q, \eventFunction(o, \pi_k(s)))}}_{=1} \sk
	&\stackrel{\text{(ii)}}{\le} \sum_{o\in \cO}\sum_{q\in \cQ} \counter_k(q,o,\pi_k(q,o)) D_{q,o}\sum_{o'\in \cO}   \Big|\widetilde P_k(o'|o,\pi_k(s)) - P(o'|o,\pi_k(s)) \Big|	
     \sk
    &\le  \sum_{a\in \cA}\sum_{o\in \cO}\sum_{q\in \cQ} \counter_k(q,o,a) D_{q,o}\sum_{o'\in \cO}   \Big|\widetilde P_k(o'|o,a) - P(o'|o,a) \Big|	\sk
    &\le  \sum_{a\in \cA}\sum_{o\in \cO}\max_q D_{q,o}  \sum_{o'\in \cO}   \Big|\widetilde P_k(o'|o,a) - P(o'|o,a) \Big|	\sum_{q\in \cQ}\counter_k(q,o,a)\, ,
	\label{eq:Ptilde_P_diff_RM_B}
	\end{align}
where in (i) we used the definition of $\cB_s$, and where (ii) follows from Lemma \ref{lem:RM_restricted_value_bound}. Observe that 	
\begin{align*}
     \sum_{o'\in \cO}\Big|\widetilde P_k(o'|o,a) &- P(o'|o,a) \Big| \le  \sum_{o'\in \cO}\Big|\widetilde P_k(o'|o,a) - \widehat P_k(o'|o,a) \Big| +  \sum_{o'\in \cO}\Big|\widehat P_k(o'|o,a) - P(o'|o,a) \Big| \\
    &\le 
\sum_{o'\in \cO}\sqrt{\frac{2\widetilde P_k(o'|o,a)(1-\widetilde P_k(o'|o,a))}{N_k(o,a)}\beta'_{N_k(o,a)}\big(\tfrac{\delta}{2O^2A}\big)} \sk
&+ \sum_{o'\in \cO}\sqrt{\frac{2P(o'|o,a)(1-P(o'|o,a))}{N_k(o,a)}\beta'_{N_k(o,a)}\big(\tfrac{\delta}{2O^2A}\big)} +  \frac{2}{3N_k(o,a)}\beta'_{N_k(o,a)}\big(\tfrac{\delta}{2O^2A}\big)
\sk
&\stackrel{\text{(i)}}{\le} \sqrt{\beta'_{T}\big(\tfrac{\delta}{2O^2A}\big)} \sum_{o'\in \cO}\sqrt{\frac{2\widehat P_k(o'|o,a)(1-\widehat P_k(o'|o,a))}{N_k(o,a)}} \sk 
&+ \sqrt{\beta'_{T}\big(\tfrac{\delta}{2O^2A}\big)} \sum_{o'\in \cO}\sqrt{\frac{2P(o'|o,a)(1-P(o'|o,a))}{N_k(o,a)}} +  \frac{4}{N_k(o,a)}\beta'_{T}\big(\tfrac{\delta}{2O^2A}\big) \sk
\label{eq:bound_on_p_diff}
&\stackrel{\text{(ii)}}{\le} \sqrt{8\beta'_{T}\big(\tfrac{\delta}{2O^2A}\big)\frac{K_{o,a}}{N_k(o,a)}}  +  \frac{4\beta'_{T}\big(\tfrac{\delta}{2O^2A}\big)}{N_k(o,a)}\,,
\end{align*}
where (i) follows from Lemma \ref{lem:SqrtVar_pqBern}, 
and where (ii) uses the fact that for a distribution $p\in \Delta_\cO$ with $K$ non-zero elements, $\sum_{o\in \cO} \sqrt{p(o)(1-p(o))} \le \sqrt{K-1}$ ---see Proof of Theorem 2 
for details. Hence,
\begin{align*}
	\mathbf\counter_k (\widetilde{\mathbf{P}}^\times_k - \mathbf{P}^\times_k) w_k = \sqrt{8\beta'_{T}\big(\tfrac{\delta}{2O^2A}\big)} \sum_{o,a}\counter_k(o,a)\sqrt{\frac{K_{o,a}}{N_k(o,a)}}\max_q D_{q,o}  +  4\Dcp \beta'_{T}\big(\tfrac{\delta}{2O^2A}\big)\sum_{o,a}\frac{\counter_k(o,a)}{N_k(o,a)}\, .
\end{align*}
The rest of the proof follows along the lines of the proof of Theorems \ref{thm:regret_UCRLSAL1_deterministic} and \ref{thm:regret_UCRLSAB}.
\ep

\section{Regret Lower Bound}\label{app:regret_LB}
In this section, we prove Theorem \ref{thm:LB}. Our proof uses the machinery of establishing a minimax regret lower bound in \cite{jaksch2010near} for tabular MDPs. (We also refer to  \citep[Chapter~38.7]{lattimore2020bandit}.) This machinery for tabular MDPs consists in crafting a worst-case MDP and showing that the regret under any algorithm on the MDP is lower bounded. 
We take a similar approach here but stress that constructing a worst-case MDPRM entails constructing a worst-case reward machine and a labeled MDP simultaneously. In terms of notations and presentation, we closely follow  \citep[Chapter~38.7]{lattimore2020bandit}. 

\begin{figure}[h]
\centering
\tiny
      \def\svgwidth{0.45\textwidth}
      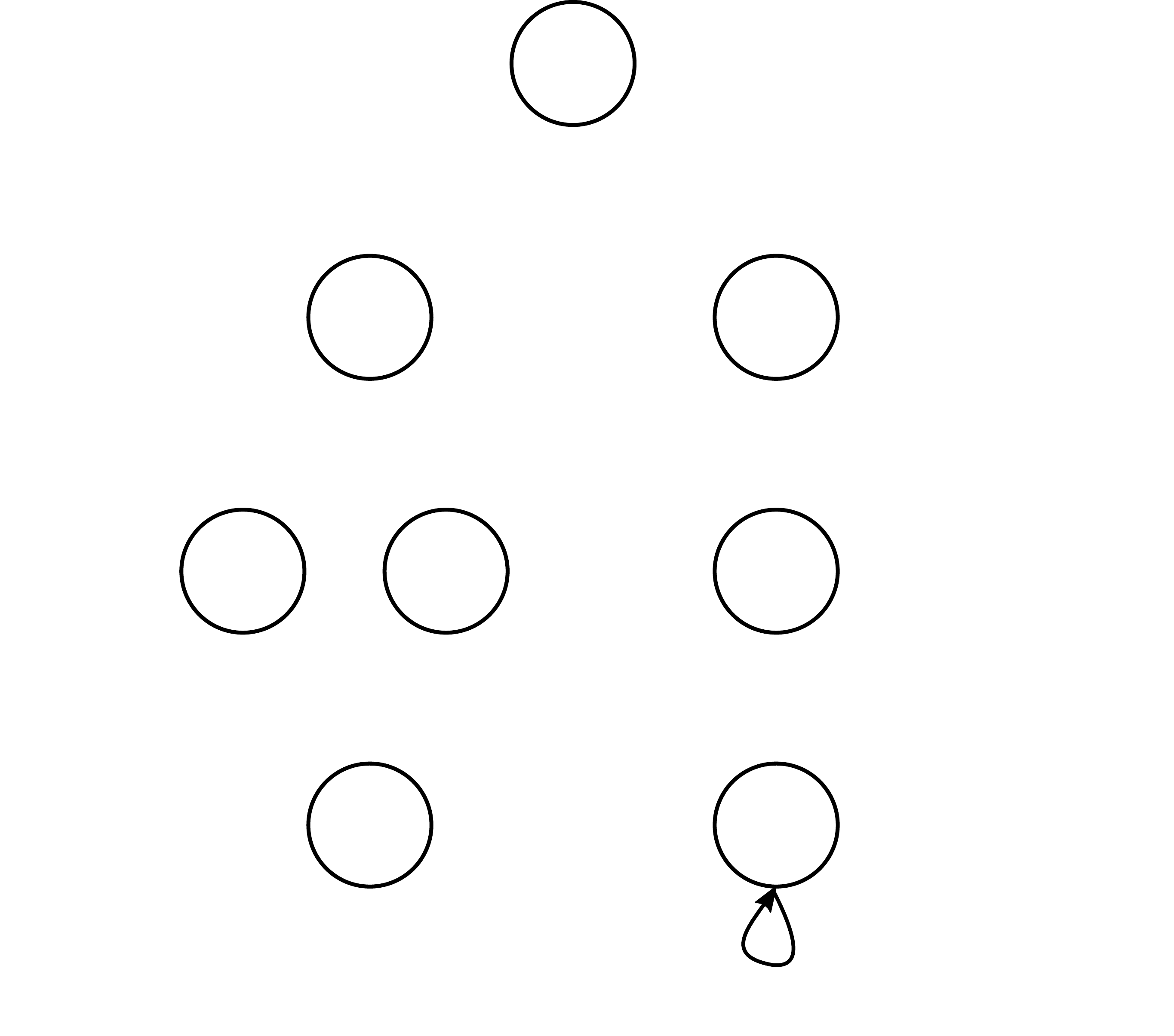
  \vspace{-1mm}
  {\caption{Construction of the underlying labeled MDP for the LB with $A = 2$ and $O = 8$, based on the worst-case MDP in \citep[Chapter~38.7]{lattimore2020bandit}.}
  \label{fig:LB_SA}}
\end{figure}

\begin{figure}[h]
\tiny
      \def\svgwidth{.78\textwidth}
      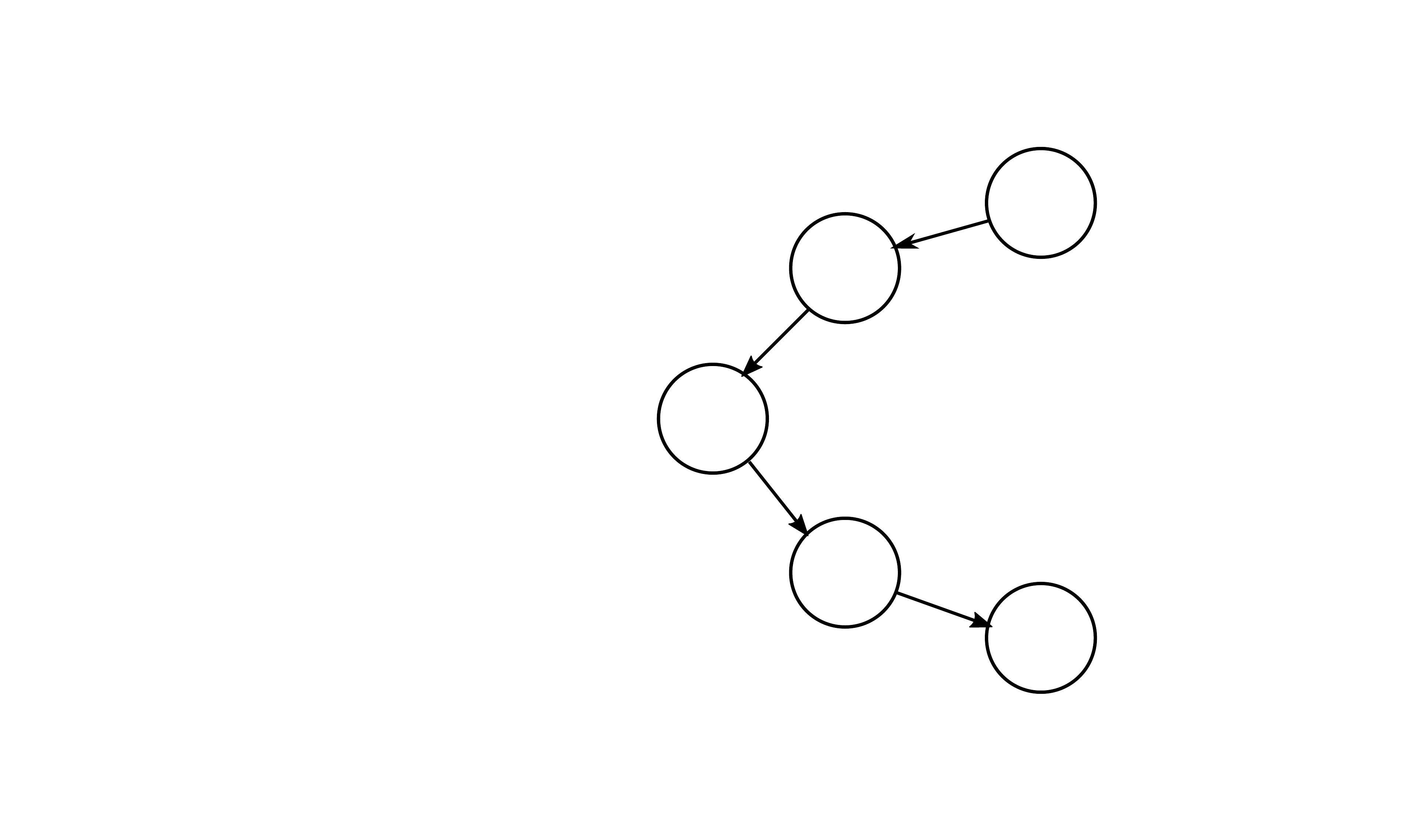
  \vspace{-3mm}
  {\caption{Construction of the underlying RM for the lower bound with a double-cyclic a `good' cycle giving rewards and `bad' cycle of similar length giving no reward.}
  \label{fig:LB_MDP}}
\end{figure}




\emph{Proof (of Theorem \ref{thm:LB}).} 
To prove the theorem, we construct a worst-case MDPRM, which can be seen as an MDPRM that models a bandit problem with approximately $OA$ arms, such that obtaining the reward requires to pick the `good arm' $Q$ times. Figures \ref{fig:LB_MDP} and \ref{fig:LB_SA} show the construction, given $O$ and $A$: We build a tree of minimum depth with at most $A$ children for each node using exactly $O - 2$ observations. The root of the tree is denoted $o_0$ and transitions within the tree are deterministic. So, in a node of the tree the agent can simply select the child to transition to. Let $L$ be the number of leaves, and let us index observations as $o_1, o_2, \dots, o_L$. The last two observations are $o_A$ and $o_B$ where events are given as detailed later. Then, for each $i \in [L]$ the agent can choose any action $a \in \cA$ and transitions to either $o_A$ or $o_B$ according to:
\als{
P(o_A|o_i, a) = \frac{1}{2} + \epsilon(a, i) \quad \textit{and} \quad P(o_B|o_i, a) = \frac{1}{2} - \epsilon(a, i),
}
where $\epsilon(a, i) = 0$ for all $(a, i)$ pairs except for one particular pair, for which $\epsilon(a, i) = \Delta>0$. ($\Delta$ will be chosen later in the proof.) The transition probabilities at $o_A$ and $o_B$ under any $a\in \cA$ satisfy:
\als{
&P(o|o, a) = 1 - \delta, \quad  P(o_0|o,a) = \delta,\quad o\in \{o_A,o_B\}\,.
}
Let us choose $\delta = \frac{6Q}{\Dcp}$. Note that by the assumptions of the theorem, $\delta\in (0,1]$. Furthermore, this choice ensures that the diameter of the cross-product  MDP associate to the described MDPRM is at most $\Dcp$, regardless of the value of $\Delta$. Also, for the diameter of the labeled MDP, $D$, we will have $D=\frac{4}{\delta}$.

The labeling function is defined as follows. Since we assume $\Sigma \geq 4$, we can consider three events $\sigma_A, \sigma_B, \sigma_{A \cap B}$ and define labeling function as follows: For all action $a\in \cA$,
\als{
&L(o_A, a, o_0) = \sigma_A, \qquad L(o_A, a, o_A) = \sigma_{A \cap B},\qquad 
L(o_B, a, o_0) = \sigma_B, \qquad L(o_B, a, o_B) = \sigma_{A \cap B}.
}
To build the RM,  we let $N = \lceil (Q - 1)/2 \rceil$ and $N' = \lfloor (Q - 1)/2 \rfloor$ so that $N+N'=Q-1$. The idea is to arrange the $Q$ many nodes of the RM into 2 cycles of lengths $N$ and $N'$; see Figure \ref{fig:LB_SA}. To this effect, we let $q_0$ be the origin. Then, the set $\{q_i\}_{i = 0}^N$ of RM states defines the `good' cycle, whereas the set $\{q'_j\}_{j = 1}^{N'}\cup \{q_0\}$ define the `bad' cycle. Then, we build the RM transition function $\tau$ and reward function $\nu$, for all $i \in [N]$ and all $j \in [N']$:
\als{
\tau(q_0, \sigma_A) = q_{1}, \qquad \qquad &\nu(q_0, \sigma_A) = 1,\\
\tau(q_0, \sigma_B) = q'_{1}, \qquad \qquad &\nu(q_0, \sigma_B) = 0,\\
\tau(q_i, \sigma_A) = q_{i + 1}, \qquad \qquad &\nu(q_i, \sigma_A) = 1,\\
\tau(q_i, \sigma_B) = q_{i + 1}, \qquad \qquad &\nu(q_i, \sigma_B) = 1,\\
\tau(q_i, \sigma_{A \cap B}) = q_{i}, \qquad \qquad &\nu(q_i, \sigma_{A \cap B}) = 1,\\
\tau(q_{N}, \sigma_A) = q_{0}, \qquad \qquad &\nu(q_{N}, \sigma_A) = 1,\\
\tau(q_{N}, \sigma_B) = q_{0}, \qquad \qquad &\nu(q_{N}, \sigma_B) = 1,\\
\tau(q_{N}, \sigma_{A \cap B}) = q_{N}, \qquad \qquad &\nu(q_{N}, \sigma_{A \cap B}) = 1,\\
\tau(q'_j, \sigma_A) = q'_{j + 1}, \qquad \qquad &\nu(q'_j, \sigma_A) = 0,\\
\tau(q'_j, \sigma_B) = q'_{j + 1}, \qquad \qquad &\nu(q'_j, \sigma_B) = 0,\\
\tau(q'_{N'}, \sigma_A) = q_{0}, \qquad \qquad &\nu(q'_{N'}, \sigma_A) = 0,\\
\tau(q'_{N'}, \sigma_B) = q_{0}, \qquad \qquad &\nu(q'_{N'}, \sigma_B) = 0,\\
}
where all non-specified transitions imply no change of state, and where all non-specified rewards are zero. This means that in $q_0$, the agent needs to realize the event $\sigma_A$ to initiate a rotation of the `good' cycle, where in all states the agent will get a reward when staying in either $o_A$ or $o_B$ and progresses one step forward in the cycle when leaving one of both RM-states. On the other hand, if the agent is in $q_0$, she receives the event $\sigma_B$ and then  initiates a rotation of the `bad' cycle, without any reward but similar length and transitions as for the `good' cycle.

In summary, each time the agent arrives in $s_0 = (o_0, q_0)$, she selects which leaf to visit and then chooses an action from that leaf. This corresponds to choosing one of $k = LA = \Omega(OA)$ meta actions. The optimal policy is to select the meta action with the largest probability of transitioning to the observation $o_A$. The choice of $\delta$ ensures that the agent expects to stay at state $o_A$ or $o_B$ for approximately $D$ rounds. Since all choices are equivalent when $q \neq q_0$, the agent expects to make about $\frac{2T}{DQ}$
decisions and the rewards are roughly in $[0, \frac{DQ}{8}]$, or $3DQ = 2\Dcp$, so we should expect the regret to be $\Omega(\Dcp\sqrt{kT/\Dcp}) = \Omega(\sqrt{T\Dcp OA}).$

\paragraph{Characterization of the MDPRM.} Using the introduced notations, we introduce: 
$\cL:= \{(q_0, o, a): a \in \cA \text{ and } o \text{ is a leaf of the tree}\}$ and $\cL^M := \{(o, a): a \in \cA \text{ and } o \text{ is a leaf of the tree}\}$. 
By definition, both have $k$ elements. Then, let $M_0$ be the MDPRM with $\epsilon(o, a) = 0$ for all pairs in $\cL^M$. Then let $M_j$ be the MDPRM with $\epsilon(o, a) = \Delta$ for the $j$-th observation-action pair in the set $\cL^M$. Similarly to \citep{lattimore2020bandit}, we define the stopping time $T_{\text{stop}}$ as the first time when the number of visits of $(q_0, s_0)$ is at least $T/\Dcp - 1$, or $T$ if the state $(q_0, s_0)$ is not visited enough: 
\als{
T_{\text{stop}} = \min \bigg\{T, \min \Big\{t: \sum_{t'=1}^t \indic{s_{t'} = (q_0, o_0)} \geq \frac{T}{\Dcp} - 1 \Big\}\bigg\}\,.
}
Also, let $T_j$ be the number of visits to the $j$-th triplet of $\cL$ until $T_{\text{stop}}$ and $T_{\text{tot}} = \sum_{j = 1}^k T_j$. We also let $P_j, 0 \leq j \leq k$ denote the probability distribution of $T_1, \dots, T_k$ induced by the interaction of $\pi$ and $M_j$ and let $\mathbb{E}_j[\cdot]$ be the expectation with respect to $P_j$.

Now, we study the characteristics of the MDPRM. In doing so, we first build upon \citep[Claim~38.9]{lattimore2020bandit} that establishes that the diameter of the underlying MDP of $M_j$, denoted by $D(M_j)$, is bounded by $D$ for all $j \in [k]$. Then, we have for $\Dcp(M_j)$ cross-product diameter of the MDPRM $M_j$:
\als{
\Dcp(M_j) \leq DN + DN \sum_{i = 1}^\infty \frac{1}{2^i} + DN' \leq \frac{3}{2}QD = \Dcp.
}
The first inequality can be interpreted as the fact that the cross-product diameter is smaller that completing the 2 loops of the RM plus accounting the probability to have a transition to the ``wrong" loop when in $q_0$. The rest follows by construction and we note that we can ignore $\Delta$ due to the fact that it can only reduce the diameters.

Following the same arguments as in Claim 38.10 of \citep{lattimore2020bandit}, there exist universal constants $0 < c_1 < c_2 < \infty$ such that $\Dcp \mathbb{E}_0[T_\sigma]/T \in [c_1, c_2]$. By construction, we have
\als{
\frac{\Dcp \E_0[T_{\text{tot}}]}{T} &\leq \frac{\E[T_{\text{tot}}]}{OA}\leq \frac{T}{DN'OA} \leq c_2, \qquad 
\frac{\Dcp \E[T_{\text{tot}}]}{T} \geq \frac{\E_0[T_{\text{tot}}]}{OA}\geq \frac{T}{DNOA} \geq c_1.
}
Finally, we write $\EE[\kR_j(T)]$ the expected regret of policy $\pi$ in the MDPRM $M_j$ over $T$ steps and prove that there exists a universal constant $c_3 > 0$ such that:
\als{
\EE[\kR_j(T)] \geq c_3 \Delta \Dcp \E[T_{\text{tot}} - T_j]\, .
}
To prove this result, we first write the definition of the expected regret:
\als{
\EE[\kR_j(T)] = \sum_{t=1}^T \E_j^\star[r_t] - \sum_{t=1}^T \E_j[r_t],
}
where $\E_j^\star$ is the expectation in MDPRM $M_j$ when following the optimal policy, which mean always choosing the $j$-th element of $\cL$ when in $(q_0, o_0)$. Now, we can decompose the cumulative reward by ``episodes", where a new episode start whenever reaching $(q_0, o_0)$. By construction and using our knowledge of the optimal policy, this yields:
\als{
\EE[\kR_j(T)] &\geq \E_j[T_{\text{tot}}]\big (\frac{1}{2} + \Delta \big )\frac{DN}{4} - \E[T_{\text{tot}} - T_j]\frac{DN}{8} - \E_j[T_j]\big (\frac{1}{2} + \Delta)\frac{DN}{4}\\
&= \E_j[T_{\text{tot}} - T_j]\Delta \frac{DN}{4},
}
or by definition of $D$ and $N$ there exists a universal constant $c_3 > 0$ such that $c_3\Dcp \geq \frac{DN}{4}$, which allows us to conclude.

\paragraph{The Final Lower Bound.} Let $\mathrm{D}(P,Q)$ denote the Kullback-Leibler divergence between two probability distributions $P$ and $Q$. Similarly to \citep[Chapter~38.7]{lattimore2020bandit} and \citep{jaksch2010near} (as well as lower bound proofs for bandit problems), we have $\mathrm D(P_0, P_j) = \E_0[T_j]d(1/2, 1/2 + \Delta)$, where $d(p, q)$ is the relative entropy between Bernoulli distributions with respective means $p$ and $q$. Now the conclusion of the proof is exactly the same as for MDPs \citep{jaksch2010near}: We assume that the chosen $\Delta$ will satisfy $\Delta \leq 1/4$, then using the entropy inequalities from \citep[Equation~14.16]{lattimore2020bandit}, we have:
\als{
\mathrm{D}(P_0, P_j) \leq 4\Delta^2 \E_0[T_j].
}
Then following the same steps as in \citep[Chapter~38.7]{lattimore2020bandit} and using Pinsker's inequality, and using the fact that $0\leq T_{\text{tot}} - T_j \leq T_{\text{tot}} \leq T/\Dcp$, we have
\als{
\E_j[T_{\text{tot}} - T_j] \geq \E[T_{\text{tot}} - T_j] - \frac{T}{\Dcp}\sqrt{\frac{\mathrm D(P_0, P_j)}{2}} \geq \E_0[T_{\text{tot}} -T_j] - \frac{T\Delta}{\Dcp}\sqrt{2\E_0[T_j]}.
}
Summing over $j$ and applying Cauchy-Schwarz give us
\als{
\sum_{j = 1}^k \E_j[T_{\text{tot}} - T_j] &\geq \sum_{j = 1}^k \E_0[T_{\text{tot}} - T_j] - \frac{T \Delta}{\Dcp}\sum_{j=1}^k\sqrt{2\E_0[T_j]}\\
&\geq (k - 1)\E_0[T_{\text{tot}}] - \frac{T\Delta}{\Dcp}\sqrt{2k\E_0[T_{\text{tot}}]}\\
&\geq \frac{c_1T(k-1)}{\Dcp} - \frac{T\Delta}{\Dcp}\sqrt{\frac{2c_2Tk}{\Dcp}}\, .
}
Now choosing $\Delta = \frac{c_1(k-1)}{2}\sqrt{\frac{\Dcp}{2c_2Tk}}$ yields
\als{
\sum_{j = 1}^k \E_j[T_{\text{tot}} - T_j]\geq \frac{c_1T(k-1)}{2k\Dcp}\, .
}
This implies that there exists $j$ such that $\E_j[T_{\text{tot}} - T_j]\geq \frac{c_1T(k-1)}{2k\Dcp}$, which leads to the final result using the previous lower bound on the regret
\als{
\EE[\kR_j(T)] \geq c_3\Dcp \Delta \E_j[T_{\text{tot}} - T_j] \geq \frac{c_1^2c_3T(k-1)^2}{4k}\sqrt{\frac{\Dcp}{2c_2Tk}} = c_0\sqrt{\Dcp OAT},
}
with $c_0 > 0$ being a universal constant.
\ep

\section{Useful Concentration Inequalities}
\label{sec:concentration_lemmas}

In this subsection, we collect a few useful concentration inequalities. They can be found in, e.g.,  \citep{maillard2019mathematics,lattimore2020bandit,dann2017unifying,bourel2020tightening}. We begin with the following definition:
\begin{definition}\label{def:subGaussian}
    A sequence $(Y_t)_{t}$ has conditionally $\sigma$-sub-Gaussian noise if	
	\begin{align*}
	\forall t, \forall \lambda\in\Real,\quad  \log \Esp[ \exp\big(\lambda (Y_t- \Esp[Y_t|\cF_{t-1}])\big) \big| \cF_{t-1}] \leq \frac{\lambda^2 \sigma^2}{2}\,,
	\end{align*}
    where $\cF_{t-1}$ denotes the $\sigma$-algebra generated by $Y_1,\ldots,Y_{t-1}$.
\end{definition}

\begin{lemma}[Time-Uniform Laplace Concentration for Sub-Gaussian Distributions]
\label{lem:time-uniform-Laplace-subGaussian}
	Let $Y_1,\ldots, Y_n$ be a sequence of $n$ i.i.d.~real-valued random variables with mean $\mu$, such that $Y_n-\mu$ is  $\sigma$-sub-Gaussian. Let
	$\widehat\mu_n= \frac{1}{n}\sum_{s=1}^n Y_s$ be the empirical mean estimate. Then, for all $\delta\in(0,1)$, it holds
	\begin{align*}
	\mathbb P\bigg(\exists n\in\Nat,\quad |\widehat\mu_n -\mu| \geq \sigma\sqrt{\frac{2}{n}\Big(1+\frac{1}{n}\Big)\ln\big(\sqrt{n+1}/\delta\big)}\bigg)& \leq \delta\,.
	\end{align*}		
\end{lemma}

The ``Laplace'' method refers to using the Laplace method of integration for optimization. We recall that random variables bounded in $[0,1]$ are $\tfrac{1}{2}$-sub-Gaussian. The following corollary is an immediate consequence of Lemma \ref{lem:time-uniform-Laplace-subGaussian}:

\begin{corollary}[Time-Uniform Azuma-Hoeffding Concentration using Laplace]\label{lem:time-uniform-Laplace-AzumaHoeffding}
Let $(X_t)_{t\ge 1}$ be a martingale difference sequence such that for all $t$, $X_t\in [a,b]$ almost surely for some $a,b\in \RR$. Then, for all $\delta\in (0,1)$, it holds
	\begin{align*}
	\mathbb P\bigg(\exists T \in\Nat: \sum_{t=1}^T  X_t \geq (b-a)\sqrt{ \tfrac{1}{2}(T+1) \log( \sqrt{T+1}/\delta)}\bigg) \leq \delta\,.
	\end{align*}
\end{corollary}

Lemma \ref{lem:time-uniform-Laplace-subGaussian} can be used to provide a time-uniform variant of Weissman's concentration inequality \citep{weissman2003inequalities} using the method of mixture (a.k.a.~the Laplace method): 

\begin{lemma}[Time-Uniform L1-Deviation Bound for Categorical Distributions Using Laplace]
\label{lem:time-uniform-Laplace-L1}
Consider a finite alphabet $\cX$ and let $P$ be a probability distribution over $\cX$. Let $(X_t)_{t\ge 1}$ be a sequence of i.i.d.~random variables distributed according to $P$, and let $\widehat P_n(x) =\frac{1}{n}\sum_{i=1}^n \indic{X_i=x}$ for all $x\in \cX$. Then, for all $\delta\in (0,1)$, 
\als{
\PP\left(\exists n\in \NN: \, \|P - \widehat{P}_n\|_1 \geq  \sqrt{\frac{2}{n}\Big(1+\frac{1}{n}\Big)\log\bigg(\sqrt{n+1}\frac{2^{|\cX|}-2}{\delta}\bigg)} \right) \leq \delta\,.
}
\end{lemma}

The following lemma provides a time-uniform Bernstein-type concentration inequality for bounded random variables: 



\begin{lemma}[Time-Uniform Bernstein for Bounded Random Variables Using Peeling]
\label{lem:Bernstein_peeling}
  Let $Z = (Z_t)_{t\in \NN}$ be a sequence of random variables generated by a predictable process, and $\cF=(\cF_t)_{t}$ be its natural filtration. Assume for all $t\in \NN$, $|Z_t|\le b$ and $\EE[Z_{s}^2|\cF_{s-1}] \le v$ for some positive numbers $v$ and $b$.
  Let $n$ be an integer-valued (and possibly unbounded) random variable that is $\cF$-measurable. Then, for all $\delta\in (0,1)$,
  \als{
\PP\bigg(\exists n\in \NN, \,\, \frac{1}{n}\sum_{t=1}^{n}Z_t
\geq \sqrt{\frac{2\ell_n(\delta) v}{n}} +  \frac{\ell_n(\delta)b}{3n} \bigg) &\leq \delta\, , 
\\
\PP\bigg(\exists n\in \NN, \,\, \frac{1}{n}\sum_{t=1}^{n}Z_t \leq -\sqrt{\frac{2\ell_n(\delta) v}{n}} -  \frac{\ell_n(\delta)b}{3n} \bigg)  \leq \delta\,  ,
}
where $\ell_{n}(\delta):= \eta\log\Big(\frac{\log(n)\log(\eta n)}{\delta \log^2(\eta)}\Big)$, with $\eta>1$ being an arbitrary parameter.
\end{lemma}

Lemma \ref{lem:Bernstein_peeling} is derived from Lemma~2.4 in \citep{maillard2019mathematics}. We note that any $\eta>1$ is valid here, but numerically optimizing the bound shows that $\eta=1.12$ seems to be a good choice and yields a small bound. For example, when $(X_t)_{t\in \NN}$ is a  sequence of i.i.d.~Bernoulli random  variables with mean $\mu$, we have, for all $\delta\in (0,1)$,
  \als{
\PP\bigg(\exists n\in \NN, \,\, \mu - \frac{1}{n}\sum_{t=1}^{n}X_t
\geq \sqrt{\frac{2\ell_n(\delta) \mu(1-\mu)}{n}} +  \frac{\ell_n(\delta)}{3n} \bigg) &\leq \delta\, , \\
\PP\bigg(\exists n\in \NN, \,\, \mu - \frac{1}{n}\sum_{t=1}^{n}X_t \leq -\sqrt{\frac{2\ell_n(\delta) \mu(1-\mu)}{n}} -  \frac{\ell_n(\delta)}{3n} \bigg)  &\leq \delta\,  ,
}

\end{document}